\documentclass[11pt]{article}
\usepackage{amsmath,amssymb,amsfonts}
\usepackage{algorithmic}
\usepackage{graphicx}
\usepackage{textcomp}
\usepackage{xcolor}
\usepackage[round,authoryear]{natbib}
    
\def\BibTeX{{\rm B\kern-.05em{\sc i\kern-.025em b}\kern-.08em
    T\kern-.1667em\lower.7ex\hbox{E}\kern-.125emX}}
        
\usepackage{fancyhdr}

\newcommand{\commentout}[1]{}
\newcommand{\newresultsout}[1]{}
\newcommand{\alloldresultsout}[1]{}
\newcommand{\allolderresultsout}[1]{}
\renewcommand{\cite}{\citep}
\usepackage{hyperref}
\usepackage[ruled]{algorithm2e}
\usepackage{bm}  %%bold math
\usepackage{mathrsfs}
\usepackage{tikz}

\newcommand{\Xb}{\mathrm{\mathbf{X}}}

\newcommand{\Yb}{\mathrm{\mathbf{Y}}}

\newcommand{\Gub}{\mathrm{\mathbf{\hat{G}}}}

\newcommand{\ab}{\mathrm{\mathbf{a}}}

\newcommand{\omeb}{\bm{\omega_e}}
\newcommand{\Omxb}{\bm{\Omega_x}}
\newcommand{\omxb}{\bm{\omega_x}}

\newcommand{\Omx}{\Omega_x}
\newcommand{\omx}{\omega_x}

\newcommand{\Ome}{\Omega_e}

\newcommand{\fub}{\mathrm{\mathbf{f}}}

\newcommand{\ConSpace}{\bm{\mathcal{Y}}}
\newcommand{\DenSpace}{\bm{\mathcal{X}}}
\newcommand{\StateSpace}{\bm{\mathcal{S}}}

\newcommand{\xb}{\mathrm{\mathbf{x}}}

\newcommand{\yb}{\mathrm{\mathbf{y}}}
\newcommand{\x}{\mathrm{x}}

\newcommand{\bact}{{\em BayesAct}\xspace}

\newcommand{\act}{{\em ACT}\xspace}

\newcommand{\epa}[3]{(EPA:$\{#1,#2,#3\}$)}

\newcommand{\actlabel}[1]{}

\newcommand{\identity}[1]{{\em #1}}

\newcommand\minus{%
  \setbox0=\hbox{-}%
  \vcenter{%
    \hrule width\wd0 height \the\fontdimen8\textfont3%
  }%
}

\makeatletter
\def\grd@save@target#1{%
  \def\grd@target{#1}}
\def\grd@save@start#1{%
  \def\grd@start{#1}}
\tikzset{
  grid with coordinates/.style={
    to path={%
      \pgfextra{%
        \edef\grd@@target{(\tikztotarget)}%
        \tikz@scan@one@point\grd@save@target\grd@@target\relax
        \edef\grd@@start{(\tikztostart)}%
        \tikz@scan@one@point\grd@save@start\grd@@start\relax
        \draw[minor help lines] (\tikztostart) grid (\tikztotarget);
        \draw[major help lines] (\tikztostart) grid (\tikztotarget);
        \grd@start
        \pgfmathsetmacro{\grd@xa}{\the\pgf@x/1cm}
        \pgfmathsetmacro{\grd@ya}{\the\pgf@y/1cm}
        \grd@target
        \pgfmathsetmacro{\grd@xb}{\the\pgf@x/1cm}
        \pgfmathsetmacro{\grd@yb}{\the\pgf@y/1cm}
        \pgfmathsetmacro{\grd@xc}{\grd@xa + \pgfkeysvalueof{/tikz/grid with coordinates/major step}}
        \pgfmathsetmacro{\grd@yc}{\grd@ya + \pgfkeysvalueof{/tikz/grid with coordinates/major step}}
        \foreach \x in {\grd@xa,\grd@xc,...,\grd@xb}
        \node[anchor=north] at (\x,\grd@ya) {\pgfmathprintnumber{\x}};
        \foreach \y in {\grd@ya,\grd@yc,...,\grd@yb}
        \node[anchor=east] at (\grd@xa,\y) {\pgfmathprintnumber{\y}};
      }
    }
  },
  minor help lines/.style={
    help lines,
    step=\pgfkeysvalueof{/tikz/grid with coordinates/minor step}
  },
  major help lines/.style={
    help lines,
    line width=\pgfkeysvalueof{/tikz/grid with coordinates/major line width},
    step=\pgfkeysvalueof{/tikz/grid with coordinates/major step}
  },
  grid with coordinates/.cd,
  minor step/.initial=.2,
  major step/.initial=1,
  major line width/.initial=2pt,
}
\makeatother

\begin{document}

%\title{Cognitive Biases as the Control of Affect and Uncertainty}
%\title{``The way we do things'':\\ Social-Psychological Control of Affect and Uncertainty}
%\title{The Social-Psychological Control of Affect and Uncertainty}
\title{``Conservatives Overfit, Liberals Underfit'':\\ The Social-Psychological Control of Affect and Uncertainty}

\author{
Jesse Hoey  \\
Cheriton School of Computer Science\\
University of Waterloo\\
Waterloo, Ontario,\\
Canada, N2L3G1  \\
\url{jhoey@cs.uwaterloo.ca} \\
\and
%Tobias Schr\"{o}der\\
%Potsdam University of Applied Sciences,\\
%Kiepenheuerallee 5,\\
%14469 Potsdam, Germany\\
%\url{post@tobiasschroeder.de}
%\and
Neil MacKinnon\\
University of Guelph\\
Guelph, Ontario, \\
Canada\\
\url{nmackinn@uoguelph.ca}\\
}

\maketitle
\thispagestyle{fancy}

\begin{abstract}
  The presence of artificial agents in human social networks is growing. From chatbots to robots, human experience in the developed world is moving towards a socio-technical system in which agents can be technological or biological, with increasingly blurred distinctions between. Given that emotion is a key element of human interaction, enabling artificial agents with the ability to reason about affect is a key stepping stone towards a future in which technological agents and humans can work together. 
  This paper presents work on building intelligent computational agents that integrate both emotion and cognition.
  These agents are grounded in the well-established social-psychological
  Bayesian Affect Control Theory (\bact).
  The core idea of \bact is that humans are motivated in their social interactions by affective alignment: they strive for their social experiences to be coherent at a deep, emotional level with their sense of identity and general world views as constructed through culturally shared symbols. This affective alignment creates cohesive bonds between group members, and is instrumental for collaborations to solidify as relational group commitments.
  \bact agents are motivated in their social interactions by a combination of affective alignment and decision theoretic reasoning, trading the two off as a function of the uncertainty or unpredictability of the situation. This paper provides a high-level view of dual process theories and advances \bact as a plausible, computationally tractable model based in social-psychological theory. We introduce a revised \bact model that more deeply integrates social-psychological theorising, and we demonstrate a key component of the model as being sufficient to account for cognitive biases about fairness, dissonance and conformity. We show how the model can unify different exploration strategies in reinforcement learning.
  %We close with ethical and philosophical discussion.
\end{abstract}

%  affect control theory, Markov decision process, free energy, active inference, social order, artificial intelligence

\section{Introduction}
A key element of human experience is emotion, and enabling artificial agents with the ability to reason about emotions is a key stepping stone towards a future in which artificial intelligence (AI) and humans can work together cooperatively in social dilemmas,\footnote{A social dilemma is a game with {\em uncompensated interdependencies} (externalities)~\cite{Kollock1998}: each person's actions in the game affect other persons without their explicit consent (e.g. without compensating them).} while respecting ethical, moral and normative orders in society.  Our vision is to build intelligent computational agents that parsimoniously integrate both emotion and cognition, that are able to become members of a socio-technical system.  We ground our vision in a social-psychological theory of affective alignment and social order called \bact~\cite{HoeyBACT15,SchroederHoeyRogers2016}, which is based on the sociological Affect Control Theory~\cite{MacKinnon1994,Heise2007}.

We pursue a pragmatic approach to understanding intelligence and building artificial intelligence (AI). In the same spirit as early AI researchers, we seek to build a mechanical entity that has the same {\em general} intelligent capacity as a human, based on theory that is (at least somewhat) grounded in social psychological research into human intelligence and being. However, we do not attempt to build a replica of the human {\em brain}, rather we seek a replica of the human {\em mind}, with information channels to the outside world that can be implemented using arbitrary sensors and actuators. Thus, while this theory will require {\em embodiment}, is it not bound to be {\em anthropomorphic}. Central to our approach is to find a {\em computationally tractable} and {\em easily interpretable} model of human intelligence, such that it can be built practically and evaluated as a valuable member of society. To satisfy the interpretability requirement, we seek {\em parsimonious} models of the link from an agent's actions to its observations, the parsimony being defined by the minimal complexity required to adequately align behaviours in social situations. This minimal complexity (most parsimonious) model is the one that ``forces'' social agents to interact constructively to solve social dilemmas and allows agents to evaluate and interact with the future by comparing their state to a model of only the optimal solution to the social dilemma, instead of to a model of all possible states~\cite{Ramstead2019}. Thus, the {\em generative model} (the world, or simply another agent in a closed dyadic situation) must be inverted to create a {\em recognition model} in the agent that mirrors it. 

Parsimonious models have previously been explored in the context of perceptual inference~\cite{Dayan1995}. In the Ecological Free Energy Principle, this has been generalized to include embodied intelligent agents capable of autonomous action~\cite{Bruineveld2019}. The core principle is that agents frequent certain configurations of the world, and value is placed on those states of the world that they frequent. \bact provides a simple, computational mechanism that brings a {\em social} element into the world, and into the agent's world model, and suggests that intelligent agents living in a social setting may use a sharing mechanism based on emotion which allows them to attend to (and thereby frequent) the social order in which they are embedded, their social econiche~\cite{Bruineveld2019}. This sharing mechanism generalizes across actions and world configurations including other social agents, constituting a highly complex environment. 

This paper makes two primary contributions. First, the exposition of a revised version of the \bact model which more coherently ties it to the underlying sociologial theorising~\cite{Heise2007}. We introduce the {\em somatic transform} as a mathematical method for linking affective and deliberative reasoning at the meta-cognitive level, and link this transform to other dual process models from cognitive neuroscience, behavioral economics, and multi-agent reinforcement learning. Second, the application of \bact to a set of three cognitive bias experiments, showing how this single idea can provide an explanation across different domains. We also connect \bact to theorising about the Bayesian brain~\cite{Friston_brain_2010}, and active inference~\cite{Friston2012}. 

We first discuss dual process models in general in Section~\ref{sec:dualprocess}, and \bact in particular in Section~\ref{sec:bact}. We show how \bact can be used to explain behavioural effects in fairness (Section~\ref{sec:bos}), dissonance (Section~\ref{sec:cogdis}) and conformity (Section~\ref{sec:conform}). In the discussion, we review two key practical application areas, one in online collaborative networks (e.g. GitHub)~\cite{hoey2018artificial} and the other in building assistive technologies for persons with dementia~\cite{RobillardHoey2018}. We briefly discuss other ethical and philosophical considerations and conclude. Technical details about \bact can be found at {\tt\url{bayesact.ca}}.

\section{Dual Process Models}
\label{sec:dualprocess}
Human and artificial intelligent agents are faced with the computational problem posed by the complexity of the social order encoded as sensory inputs. Agents must find a way to map this high dimensional input space into an equally high dimensional action space. Agents can handle the complexity of the input by constructing a representation of it, and then preforming calculations over this representation. We will call this the {\em denotative} representation, and it is an abstraction of the physics of environment. For example, it is able to represent the positions of pieces on a chess board and make predictions about how a game will turn out given a sequence of moves, or it can represent the bids in a negotation. The denotative representation is assumed to be symbolic, but can be implemented sub-symbolically.\footnote{\label{foot:subsym} By subsymbolic, we mean as in a neural network, where the ``symbols'' are weights on neurons, and therefore somewhat difficult to interpret. However, we do not rule out the possibility that a subsymbolic representation could be used to model a symbolic one. For example in a deep reinforcement learning problem, the symbols are the actions being predicted (in fact, the values of these actions), while the neural network simply provides the mapping function.} Regardless of representation, denotative calculations rapidly become intractable, and are exacerbated by the inclusion of other intelligent agents~\cite{FeldmanHall2019}.

%%%FeldmanHall review
FeldmanHall and Shenhav~\cite{FeldmanHall2019} describe an abstract Bayesian model of the management of uncertainty. In their model, humans are motivated to reduce uncertainty in their distributions over their own actions, conditioned on their current appraisal of their state. The ability to reduce uncertainty has clear evolutionary advantages, meaning humans have an intrinsic motivation to reduce it. Affect acts as an easily accesible measure of uncertainty over actions for the agent, and provides an error signal that can be used in a feedback loop to reduce uncertainty. As uncertainty grows, e.g. in a social situation, negative affect is created and an agent is motivated to reduce this, restoring positive affect, by taking actions that reduce uncertainty. In social situations, the state of the world includes the states of other agents, such as their traits, goals and emotions. A continuum of strategies, from ``automatic'' control based on stereotypes and impressions, to ``controlled'' processes based on perspective-taking and effortful search, are used to reduce the uncertainty. This continuum maps to a Bayesian tradeoff between prior and evidence, and is well described in the \bact model by a dual-process system we explain in Section~\ref{sec:bact}.

\subsection{Complexity}
\label{sec:complexity}
As the complexity of the environment increases, %(e.g. with the addition of more intelligent agents),
an agent with fixed computational resources runs into a bound that prevents it from modeling the added complexity.
As this bound is reached, the predictions made by the denotative representation have increasing trouble matching the evidence from the world, resulting in more dispersed estimates of denotative state and an impoverished mechanism for predicting the future. 
%This resource bound results in a more dispersed estimates of the denotative state representation,  %. The reason that it is more dispersed is
%because the predictions it makes 
%made by thae agent's (limited) internal denotative representation
%have increasing trouble matching the evidence from the world.
%While mismatches can be ``averaged'' in some way, the average itself is not necessarily any better than the original prediction.
Such an agent can handle this inability to predict the future by believing the predictions and ignoring the evidence (underfitting, leading to high bias, low variance behaviour), or by relaxing the predictions and believing the evidence (overfitting, leading to low bias, high variance behaviour). While the first agent will have difficulty adapting to change (but can see it coming), the second will have difficulty predicting change (but can adapt to it). We will denote the first type of agents (those that underfit) as ``L'' agents, and the second type (those that overfit) as ``C'' agents.
%% L agents and C agents 

An agent with a hierarchical model can do both, however, as it can create and track a lower-dimensional version of the denotative state that allows it to continue making predictions, albeit with reduced precision. In machine learning, the basic idea of approximating a complex function with another, simpler, function is referred to as variational, and if the functions are probabilistic, then variational Bayesian. An intelligent agent that is using a variational optimization technique to obtain policies of behaviour can be seen as an instance of active inference~\cite{Friston2012}. Active Inference proposes consistency between an agent's internal model and the environment in which it is embedded as a fundamental principle underlying biological agents. %this consistency is part of the total {\em free energy} of the system~\cite{Friston_brain_2010}, and has been used to gain deeper insights into inference, reinforcement learning and emotion~\cite{Joffily2013}.
The complexity of the agent's environment, defined as the total number of configurations accessible to the agent, is the true {\em free energy}. As an agent's world becomes more complex (e.g. with the addition of other social agents), this true free energy becomes intractable to model within an agent's resource bound, and so the agent must approximate. An agent's {\em variational free energy} is an internal measure of how well its (approximate) model fits the real world. The idea of minimizing this variational free energy is the same as moving towards a state of true free energy which is minimal and is consistent with the econiche in which it is embedded~\cite{Bruineveld2019,Friston_brain_2010}. Agents with better matching models avoid surprise and are better survivors. 
%% CHECK moved from uncertainty and affect - to be integrated with the paragraph above
%The way in which connotative and denotative reasoning and action selection are trade-offs in \bact is a reflection enactive inference~\cite{Friston_brain_2010}.
%Such a viewpoint treats the mind as operating to improve the match between internal model and external environment.
%Improving this match is equivalent to decreasing the total number of configurations modeled, also known as the “variational free energy”. At the limit, the variational free energy is the same as the true free energy. % (as given by the partition function).
\bact proposes the connotative space as performing a (variational) approximation to the denotative space. This approximation is necessary because of the impossibility of finding a good match at the denotative level alone, and more so in social environments which are inherently harder to predict, and are thefore less {\em valid}, or less predictable, or more ambiguous and uncertain~\cite{KahnemanKlein2009}. In less valid situations, the distribution over denotative states is more dispersed or has higher entropy. People perform more poorly than algorithms in situations of low validity~\cite{KahnemanKlein2009} as they are resorting to cognitive biases.
%quote is actually: ``The correct conclusion is that people perform significantly more porly than algorithms {\em in low validity environments}.''
%  ~\cite{FeldmanHall2019}). %The inferences described in this paper can be formulated as connotative space as a variational approximation to the denotative. % in fact, Y are parameters for the dynamics of X.

Related views of emotion include the identification of negative valence with increased uncertainty~\cite{FeldmanHall2019} or with change in free energy~\cite{Joffily2013}, or expected free energy~\cite{Hesp2018}.  One component of identity (esteem) is used to modulate a reward function in~\cite{Moutoussis2014} in order to make cooperation the more salient policy in a social dilemma. However, in~\cite{Moutoussis2014b}, interpersonal relationships in general are linked to heuristics that facilitate approximate Bayesian inference, allowing agents to circumvent the intractability of optimal Bayesian inference. Importantly, the models of self and other exist specifically to influence agent actions.
In a similar vein, \cite{Jaques2019} consider  social {\em influence} as an additional factor in an agent's utility function in a sequential social dilemma, and the relative weights of this factor can be learned by an agent about other agents. Influence is a signal of power, and we can therefore see the same concept being applied in this multi-agent reinforcement learning context. Note that in earlier work~\cite{Ray2008}, the additional factor in the utility function is parameterized by {\em envy} and {\em guilt}, which map to {\em both} power and evaluation dimensions~\footnote{Envy is a combined emotion of sadness and anger, which therefore are mixed between negativity and dominance (agression), while Guilt is a combination of joy and fear, mixing positivity with submission~\cite{Plutchik1980}.}. Importantly, the precision of the estimate of the other's type is critically important~\cite{Moutoussis2014b}, especially as ambiguity grows~\cite{FeldmanHall2019}. In a trust game, cooperative solutions are linked to valence (self-esteem) in~\cite{Moutoussis2014}, much like the tendency towards fair solutions in uncertain situations in~\cite{Bos2001} (see also Section~\ref{sec:bos} and~\ref{sec:durkheim}).
%
%envious, -0.64, -0.08, 0.12, -1.85, -0.87, -0.98, 10 100000000 000
%angry, -1.45, -0.30, 1.13, -1.77, -0.70, 1.34, 10 100000000 000
%sad, -1.88, -1.66, -2.06, -2.22, -1.07, -1.82, 10 100000000 000
%
%joyful, 2.43, 1.97, 1.33, 3.25, 2.18, 2.34, 10 100000000 000
%fearful, -1.64, -0.94, -1.15, -1.63, -1.38, -0.27, 10 100000000 000
%contrite, -0.16, -0.56, -0.33, -0.17, -0.44, -0.69, 10 100000000 000

%% last paragraph moved from Uncertianty and affect 
These approaches show how valence (and arousal in~\cite{Hesp2018} and power in~\cite{Jaques2019,Ray2008}) may be related to uncertainty (more precisely to the precision of policies). Affective responses are proposed as resulting from active inference in the conceptual model of~\cite{Smith2019}. \bact shows how this relationship can be linked to social psychological theory, providing a bridge to sociological analytics, by associating dimensions of sentiment with affective response representation in sentiments of {\em happy} vs {\em sad} (Evaluation) and of {\em angry} vs. {\em afraid} (Potency). However, the sentiment dynamics of \bact exist at the affective response representation and conscious acccess levels in the three-factor active inference model of~\cite{Smith2019}, and would be involved in conscious and subconscious interpretations in working memory~\cite{SmithKillgoreLane2018}. In fact, in the precursor model of Lane~\cite{Lane2000}, the level 5 theory is ``yet to be formulated'' (p362), and ``would involve social cognition and would focus on how differentiated awareness of self and other influences social behaviour...'' (p362). More recent work has noted that  ``the theory of mind abilities associated with the fifth level of emotional awareness'' are not yet fully addressed~\cite{Smith2019}.
We suggest \bact exists at this level, which can handle ``blends of blends of emotions''. For example, a feeling of happiness and sadness generated by thinking about a lucky friend feeling happy and proud, but a bit worried about his effect on me.

\subsection{\bact}
\label{sec:bayesact}
\bact is a computational dual process (hierarchical) model of intelligence, in which one process is continuous in one or more dimensions and is equated with human sentiment, while the other process is discrete or continuous and models human deliberative reasoning and decision making~\cite{HoeyBACT15,SchroederHoeyRogers2016}.  The dual process is built to handle uncertainty and surprise, naturally shifting between higher bias (lower variance) models in the sentiment space in more (denotatively) uncertain (invalid/unpredictable) situations, to higher precision (lower bias) models in the deliberative space in more certain (valid/predictable) environments. \bact agents will naturally shift in response to the unpredictability of their environment, but may do so starting from different individual ``set points'', defined by the parametrization of the \bact model. These differences in the management of uncertainty have been noted before in work demonstrating individual differences in a bias-variance tradeoff at the perceptual level~\cite{Glaze2018}. In keeping with a hierarchical Bayesian model of the mind, we show here how this tradeoff can exist at the level of conscious cognitive constructs about the self and identity. We return to these individual differences in Section~\ref{sec:durkheim}.

In \bact, we avoid the terms ``cognitive'' and ``emotional'', and refer instead to a {\em ``denotative''} representation and a {\em ``connotative''} one. The ``connotative'' is also referred to as ``sentiment'' or ``feeling'', both of which are ``affective'', while ``emotion'' refers specifically to a signaling mechanism described in Section~\ref{sec:bact}.
%The denotative representation is symbolic and often discrete (e.g. positions of chess pieces), but sometimes continuous (e.g. the laws of physics).
The denotative representation requires deliberative reasoning, in which sequences of futures are examined in memory to allow for selection of appropriate actions in the present. The connotative representation, on the other hand,
%is continuous and includes a measure of how discrepant the meanings of things are in context and out of context. Connotative states
is the {\em meaning} of the world at the level of sentiments or feelings in a relatively low dimensional space, and produces 
%somatic markers, extracted from memory, the indicate
indications of social (in)consistency or (in)coherence.  %in the world as far as the meanings of things go.
%This incoherence is a vector quantity in a relatively low dimensional ``feeling'' space.
Direct evidence of these meanings is obtained through emotional signals from other agents. Importantly, the consistency encoded in sentiments extends to agent actions and provides a rough (heuristic) guide over policies. The social intelligence provided by this consistency is shared by agents in a community, and motivates them to want to do things according to the same practice (``habitus''~\cite{Bourdieu1990}) which encodes the ``way we do things''. This shared practice is an approximation built to handle and alleviate computational complexity of the social world. %Thus, connotative and denotative states are inextricable because one can be recovered from the other at any time.
In \bact, any denotative state can be mapped into the same connotative space, allowing for comparisons between actions and identities, for example. %However, the two are complementary in that they describe the same common and deeper underlying reality and are both are necessary to fully understand this deeper reality.
%Clearly, the connotative state will be of limited usefulness on its own; it needs to be translated into something concrete in the world, and in particular into concrete motor movements (behaviors). Perhaps less obvious, the denotative state by itself will also be of limited usefulness due to the computational difficulties it presents as environments grow less valid~\cite{FeldmanHall2019}.
The connotative state is required to guide an agent towards socially acceptable choices of behavior that can ensure more globally optimal solutions to social dilemmas.
Therefore, \bact fits the definition of an {\em enactive} inference engine, in that it explicitly has the engagement of the agent with the world as a generative model (the mapping from actions to observations, which is unknown to, but partially controllable by, the agent)~\cite{Ramstead2019}. One can view the generative model in this case as a mirror image in another agent of the dynamics of sentiment. Once these dynamics sync up between agents, they can be used to create the inverse recognition model in which other agent's actions are more predictable. Note that this is not the same as modifying the utility function directly, but could be viewed as a modification to {\em expected utility}, a different form of reward shaping~\cite{Mataric1994}.

%  We show how a multi-agent system in which each agent uses such a dual process model becomes more efficient at learning and maintaining policies %(which are communicated through behaviour to other that induce collaboration and therefore optimal long-term solutions to social dilemmas. In this paper, we present a free energy formulation of such an abstract dual-process model, and show how variational message passing can be used to perform inference over the model. Importantly, this inference is also over {\em policies}, and is therefore solving the optimal planning problem simultaneously. We demonstrate how the sentiment model naturally takes on the role of a variational approximation to the deliberative model, allowing agents to more efficiently plan by considering fewer alternatives. We then argue that this variational approximation is communicated between agents using signals of emotion, and that these communications serve to bias humans towards convergence on a common such approximation. The commonality of this hierarchical variational approximation is what nudges agents to cooperate.
%, and to combine this with more focussed denotative state tracking. This ``approximate'' version is easily modeled as a continuous function mapping a vector in a multi-dimensional ``sentiment'' space across time. This function will denote the {\em connotative} state, as it represents the emotional meanings of things.

Consider a demonstrative simple example in which people and behaviours are characterised as either ``good'' or ``bad''. Cultural consensus is that good people will do good things, and so % If this consensus is encoded in a two dimensional sentiment space on a pair of valence axes (going from bad to good for behaviours and for identities), then
the denotative state does not have to model the situation in which good people are doing bad things, and thus can be simpler. The connotative state can therefore be linked to the denotative state with some energy functional dependent on the discrepancy between the meanings of things out of context and in context. If ``good'' people and ``good'' behaviours (out of context) are expected to be found together (in context), then inconsistency (good people doing bad things) will be surpising, will cause increased dispersion in estimates at the connotative level, and will push reasoning into the denotative level for analysis and re-labeling of actors and behaviours (maybe this is not, actually a ``good'' person, or maybe the behaviour is not a ``bad'' behaviour?). Critically, it is the sharing of these energy functionals that is required to gain this efficiency.
%Thus the problem of re-identification in ACT falls out naturally from the \bact framework.
%EXPLAIN WHY
%Thus a good person is expected to do good things to a good person, and if that same good person does something bad to 
%% add something further here.
%To see why this is so, consider a situation in which all sensations are 
%that there must be some functional
A slightly more realistic example is one in which a left (or right) leaning newspaper only has to write articles for their readership, and so can be more simply represented in their most elemental form as the emotional ``form'' of the news article.
%% CHECK: think of better example? 

The link in \bact between denotative and connotative induces a natural (Bayesian) tradeoff due to relative uncertainty. As the environment becomes less valid, %(less predictable or more uncertain~\cite{KahnemanKlein2009},
so the distribution over denotative states is more dispersed or has higher entropy), then the posterior will be more heavily influenced by the prior in the connotative state. Agents in less valid (less predictable) environments will put more weight on the connotative representation: they will make inferences and choose actions that are more in line with connotative (socio-cultural) expectations. In more valid environments, a lower entropy denotative distribution dominates the posterior. %, leading to a posterior that is more heavily influenced by the denotative state.
Agents in more valid environments will thus act more in line with denotative states and predictive dynamics, and so will be information seekers and utilizers. In a social dilemma, for example, one would expect the agents in less valid environments to cooperate (act according to social prescriptions), while agents in more valid environments will defect (act decision theoretically rationally). This is in line with experiments showing how humans tend to rely more on fairness in uncertain social sitautions~\cite{Bos2001}, and act more pro-socially (cooperate in a public goods game) in ambiguous situations (ones in which risk is hard to evaluate, see~\cite{VivesFeldmanHall208}). In \bact, risk is represented by the transition dynamics parameters in the denotative space. If the distribution over these parameters has lower entropy, then risk is more well defined, and so ambiguity (the uncertainty in risk) is lower.

%\subsection{Connotative Variational}
\subsection{Active Inference}
In \bact, we associate the connotative state as a variational approximation to the denotative state, and identify it as a mechanism for encoding efficient policies in the environment that includes a (potential) social group. Agents minimizing their free energy using such a dual-process model will engage in active inference and will learn a model of emotional dynamics that can ease the computational load on the denotative representation. Further, if the variational approximations are linked across agents, then the resulting social group will learn to {\em share} the same approximation in order to more efficiently solve social dilemmas. It is precisely because the collaborative solution to the dilemma yields higher payouts for all individuals that the connotative representation is selected because it is consistent with that of other agents. A secondary, normally multi-modal, signaling mechanism is used to facilitate this linkage across agents. These signals are termed ``emotional'' and ensure that the agents' connotative spaces are directly linked. This ``emotion language''~\cite{Turner2015}, communicated in part through facial expressions and paralinguistics, 
%Emotion is therefore a direct by-product of having to interact with other intelligent agents.
%As the environment contains other agents with the ability to perform complex low-entropy actions in relation to their world, the complexity of a denotative representation becomes intractable. However, a connotative, parallel, set of
allows agents to communicate what aspects of the denotative state are worth more fully exploring. As a result, the connotative state acts as a ``flashlight'' that illuminates the {\em same} part of the denotative state for all agents who share it. Such ``splotlight'' methaphors have been deeply explored in the context of psychological (usually visual) attention~\cite{Crick84}. Once this linking occurs, then the connotative state and dynamics encodes a {\em social contract} with the social group in which the agent finds itself. 

%Note that this connotative space is not synonymous with core affect as it occurs at a {\em higher level} in the brain ...

\subsection{The Relation between Cognition and Affect}
\label{sec:neil}
%%% INSERT NEIL MATERIAL HERE
%Material from Neil's chapter in here, possibly merged with the next section
As discussed in Section~\ref{sec:bayesact}, we employ the terms {\em denotative} and {\em connotative} in \bact to distinguish the cognitive meaning and representations of objects and events in high dimensional space from their affective meaning and representations in low-dimensional space. Denotative representations are associated with cognitive processing and deliberative reasoning; connotative representations with affective processing and detection of consistency between culturally-based expectations and actual occurrences. Because the distinction between cognition and affect or denotative and connotative meaning is such an essential ingredient of \bact, we discuss it in some detail before proceeding.

This distinction has been a contentious and largely unresolved issue for many years in both the neurobiological and social sciences. Historically, it can be traced back to at least the James-Lange theory of emotion~\cite{LangeJames22}, which proposed that what we feel as emotion is simply our cognitive perception of somatic states (e.g., trembling, crying, and so on) that has already occurred in direct response to external stimuli. Effectively challenged by~\citet{Cannon1929}, on the contention that bodily responses are not rapid enough to account for the immediate perception of emotional experience, interest in the James-Lange theory waned until the 1980s when it was reignited by~\citet{Zajonc1980,Zajonc1984}. On the basis of experimental evidence suggesting that subjects could make affective preferences among stimuli presented below the threshold of cognitive awareness, Zajonc concluded that ``affect and cognition are separate and partially independent systems''~\citep[p.117]{Zajonc1984} and that affect can occur without prior cognitive processing and can even precede cognition in a behavioral sequence. These conclusions set off a heated exchange with Lazarus in the 1980s that became known as the {\em primacy of cognition versus affect debate} (see~\cite{MacKinnon1994} for an extensive discussion).

The distinction between cognition and affect is even more problematic at the neurobiological level of the brain, where neuroimaging studies have failed to locate the psychological functions of cognition and affect in specific neuroanatomical regions and networks of the brain.
%In an article provocatively entitled ``Just because you're imaging the brain doesn't mean you can stop using your head,'' for example,
\citet{Cacioppo2003} suggest caution in the use of fMRI (functional magnetic resonance imaging) to study localizations of emotion and cognitive functions in the brain.
%In fact, they call the ``intuitively appealing notion that the organization of cognitive phenomena maps in a 1:1 fashion into the organization of the underlying neural substrates'' a ``category error'' in cognitive neuroscience \citeyearpar[p.654]{Cacioppo2003}, warning against equating ``the beauty of a brain image'' with ``its psychological significance'' \citeyearpar[p.657]{Cacioppo2003}.
In a similar vein, \citet{Davidson2003} includes in a list of ``seven sins'' to avoid in the study of emotion the belief that affect and cognition involve independent and separate neural circuitry and the related belief that affect is mostly subcortical and cognition mostly cortical.
%Countering these ``sins,'' he asserts ``that emotion is \ldots best understood \ldots as a set of differentiated subcomponents that are instantiated in a distributed network of cortical and subcortical circuits'' \citeyearpar[p.129]{Davidson2003}.
As pointed out by \citet{BarrettSatpute2013} in a meta-analysis of neuroimaging studies, the problem with trying to connect {\em psychological} functions to specific {\em neuroanatomical} structures or networks lies in the fact that large-scale intrinsic networks (e.g., Salience, Default Mode) are {\em domain-general} information processing networks that spill across psychological functions. On this basis one should not expect anywhere near a one-to-one correspondence of neuroanatomical structures and networks with even broad psychological categories such as levels of information processing~\cite{Ortony2005} or hierarchical levels of  emotional experience~\cite{Lane2000,SmithKillgoreLane2018}.
Many other authors have come to similar conclusions, e.g., \cite{Damasio1994,LeDoux1996,LeDouxBrown2017,Pessoa2008,Pessoa2018,DuncanBarrett2007,Franks2006,Turner2009}.
%% reduce this colossal list

%Inspired by the development of a science of large-scale, widely distributed, intrinsic brain networks (e.g., Salience, Default Mode), Barrett and Satpute's \citeyearpar{BarrettSatpute2013} meta-analysis of neuroimaging studies, strongly refutes the long-standing assumption that ``emotional, social, and cognitive phenomena are realized in the operations of separate brain regions or brain networks'' \citeyearpar[p.]{BarrettSatpute2013}. On the strength of their disconfirming findings, they conclude that ``the distinction between social, affect, and cognitive neuroscience is artificial. There is no 'affective' brain, 'social' brain, or 'cognitive' brain. Each human has one brain'' \citeyearpar[p.]{BarrettSatpute2013}. Thus,

Because \bact is a model of the human {\em mind}, not the human {\em brain}, it allows for the possibility of a distinction between cognition and affect. Our view of the relation between cognition and affect at this level consists of a balance of two ideas. On the one hand, we maintain that cognition and affect are not {\em completely} independent constituents or processes of the mind because all cognitions evoke affective feelings, if only those mild feelings generated from the perception or recognition of objects. On the other hand, to the extent cognition and affect {\em can} be distinguished as {\em partially} independent constituents or processes of the mind, we maintain that the distinction is an analytically and empirically valid one. This is most evident at the extremes of ``cold'' cognitions where the intensity of affective arousal is low and ``hot'' cognitions where arousal is quite pronounced; or, alternatively, at the extremes of affective experience largely unmediated by cognitive processing and that involving a high level of cognitive appraisal and reflection. And to the extent that cognition and affect are at least partially independent, we maintain that both are required for an adequate understanding of the human mind. 

This view of the relation between cognition and affect can be summarized as two principles~\cite{MacKinnon1994}: (1) the {\em principle of inextricability} proposes that cognition and affect are not completely independent constituents or processes of the mind, but rather a matter of relative preponderance, a continuum wherein a representation in the mind at any given moment can be predominantly cognitive or predominantly affective or anywhere in between; and (2) the {\em principle of complementarity} proposes that, as overlapping constituents or interdependent systems of phenomenological experience, both cognition and affect are necessary to understand  the human mind. While the principle of inextricability is an {\em ontological} statement about the reality of the human mind as currently understood, the principle of complementarity is an {\em epistemological} implication of this ontological view. Proposed by \citet{Bohr1950} in the 1920s to explain the contradictory images evoked by the wave-particle duality in subatomic phenomena, \citet{James1890} had developed the complementarity principle much earlier to reconcile, in his terms, the {\em substantive} and {\em transitive} parts of thought (see \cite{Stephenson1986a,Stephenson1986b}), which parallel the distinction between the denotative-cognitive and connotative-affective meanings of concepts employed in \bact. Many other authors have pointed to complementarity, including LeDoux, who opines that  ``emotion and cognition are best thought of as {\em separate but interacting} mental functions mediated by separate but interacting brain systems'' \cite[p.69, emphasis added]{LeDoux1996}, and \citet{Pessoa2008,Pessoa2018}, who focuses on brain systems underlying the interaction between emotion and cognitive processing, although he anticipates moving beyond interaction to understanding their integration in the brain. \citet{CloreOrtony2000} also argue for a mutual relation between the denotative-cognitive and connotative-affective systems of meaning, as do many others e.g. \cite{StorbeckClore2007,DuncanBarrett2007,Franks1989}. The principles of inextricability and complementarity are brought together in Franks' statement that a satisfactory resolution to the conflict between emotion and cognition ``will depend on describing how they can be inextricably linked [{\em principle of inextricability}] while capable of being in tension'' [{\em principle of complementarity}] \cite[p.55]{Franks2006}.

Assuming that the distinction between cognition and affect is valid at the psychological level of the mind, their temporal priority and causal primacy becomes a moot point if one supposes a reciprocal relationship between them. Widely suggested in the literature (e.g., \cite{Mook1987,Lazarus1984,Forgas2008,Turner2009}, this is a core assumption of affect control theory (ACT), which formalizes the reciprocal relation between cognition and affect by capitalizing on the distinction between the {\em denotative} ({\em cognitive}) and {\em connotative} ({\em affective}) meaning of words for objects and events established by Osgood and associates~\cite{Osgood1957,Osgood1969,Osgood1975}. Comprising feelings of evaluation, potency, and activity (EPA), the dimensional simplicity of connotative-affective meaning provides a portal into the dimensionally-complex denotative-cognitive representations of the world around us. Affective reactions to external objects and stimuli become ``the means by which information about the external world is translated into an internal code or representation that can be used to safely navigate the world''~\cite[p.1186]{DuncanBarrett2007}.
%Or as \cite[p.52]{CloreOrtony2000} have put it ``feelings are one of the ways in which we can represent the affective attributes of the psychological meaning of things; we can feel goodness-badness, strength-weakness, and activity-passivity. We resonate to the emotional and connotative meaning of situations by being moved ourselves.'' (See also \cite[p.338]{ClorePappas2007}; \cite[p.39]{Franks2006}; \cite[p.1199]{Russell2003} for similar expressions of this idea).
\bact moves this relation between cognition and affect to a significantly higher level by specifying a formal mathematical model that enables one to move back and forth between the denotative and connotative meanings and representations of entities.
 
\subsection{Other Dual Process Theories}
Dual process theories are well studied in social psychology~\cite{ChickenTrope1999}, but many different terms are used to refer to the two levels of processing. ``Cognitive'' processing is often referred to as deliberative, reflective~\cite{Ortony2005}, conscious~\cite{Smith2019} or ``System 2''~\cite{StanovichWest2000}, whereas ``emotional'' processing as automatic, routine~\cite{Ortony2005}, or ``System 1''~\cite{StanovichWest2000}. In many dual process theories (e.g.~\cite{FeldmanHall2019}), both deliberative and automatic systems are modeled denotatively in a constraint satisfaction network. In \bact, the connotative level is {\em affective} and serves as a low-dimensional approximation to a denotative representation. However, the connotative representation is not at the level of ``primary'' or reactive emotions (e.g. reflexes), or of core affect~\cite{Barrett2017}, but rather at the level of routine or reflective interpretations of emotions linked to procedural memory~\cite{Ortony2005}.

Behavioural economists have also tackled emotional human motivations, usually by proposing
%tackled this problem by proposing a variety of mechanisms that explain the experimental evidence of prosocial (e.g. cooperative) behaviour in humans. Early work on motivational choice~\cite{MessickMcClintock1968} proposed a probabilistic relationship between game outcomes (payoffs) and cooperative behaviour. This led to the proposition
that humans make choices based on a modified utility function that includes some reward for fairness~\cite{Rabin1993} or penalty for inequity~\cite{FehrSchmidt1999} or conformity~\cite{MasMoretti2009}. %More recently, cooperative behaviour has been linked to altruism through factors like kinship, direct reciprocity, or indirect reciprocity via reputation~\cite{Nowak2006}.
However, heuristic adjustments may not be comprehensive enough to account for human behaviour across all situations, and a morality concept that is not based on outcomes can be used as a more parsimonious account~\cite{CapraroRand2018}. The question of how this morality is defined is left open. %as an open question.

The \bact approach is to modify the {\em expected} utility by focussing computational attention on those solutions predictable by the connotative dynamics.
Note that this is a different concept than Simon's bounded rationality (SBR)~\cite{Simon67}. In SBR, the agent first performs an analysis at the symbolic level (denotative), and then “freezes” this analysis into a second denotative space called habits and coping. In ACT and in \bact, the agent gathers a fast impression and then makes predictions in an emotional space with a simple predictive function which can rapidly generate somewhat (socially) relevant predictions about future outcomes involving other agents.  In \bact, we see an emergent bounded rationality defined by uncertainty over outcomes. As the future becomes more uncertain, an emotional system automatically and softly kicks in to take up the slack. The subsequent diminishment of uncertainty is transmitted socially, shared between agents in a group.

The sociology of culture makes a distinction between heuristic cognitive biases (in so-called ``toolkit'' theories) and deeply ingrained patterns of behaviour (in so-called ``practice'' theories)~\cite{LizardoStrand2009}. Toolkit approaches have found success in explaining social structures~\cite{Martin2009}, and are hypothesised to arise from the scaffolding of the environmental and social structure The scaffolding is so complex that humans learn heuristics and tricks to get by, but the tricks are ``defined'' by the scaffolding, since they are created in order to handle exactly it. Thus, from a distance, the social structures may look complex, but in reality each individual is following a set of simple rules, like Simon's ``ant on a beach''~\cite{Martin2010}. Practice theory has been notoriously more difficult to apply, primarly because of a lack of operationalization~\cite{Li2009}. In \bact, behaviours explained by toolkit and practice theories are reflections of the same underlying affective dynamics (impression formation and the somatic transform).

%Dual process models have a long history in social
%add ref to Dual Process Theories in Social Psychology -- bottom up is denotative (denoting), top down is connotative (meaning).
%% Automatic-Reflective   Emotional-Deliberative point to Nudge
\subsection{Reinforcement Learning}
Connections of emotions with reinforcement learning have been explored by a number of authors~\cite{Moerland2017,Hogewoning07,Marinier08,ElNasr2000,Broekens2015}. In traditional reinforcement learning (RL), an agent is tasked with both learning about his world (primarily the utility of situations), and of acting in the same world.  The basic quandary for the agent is whether he should {\em exploit} his current knowledge of the world, or {\em explore} something new in the hopes of discovering even better situations: a difficult tradeoff indeed between a sure bet and a random chance. Many RL agents tackle this tradeoff using some form of {\em optimism under uncertainty}: if something has not been tried (or has not been tried for some time, or insufficiently many times), assume it will lead to high utility outcomes~\cite{BrafmanRMAX2002,Kocsis:UCT:ECML2006}.  This method works in practice because it ensures sufficient exploration, and agents don't get ``stuck''.  In traditional RL, exploitation is seen as a cognitive/rational skill requiring (usually intense) computation, since it involves predicting the future based on learned knowledge, and analyzing the costs and benefits of different strategies.  Exploration, on the other hand, is seen as something that could be guided by any number of (possibly affective) elements. For example, in~\cite{Hogewoning07}, higher valence (which is equivalent to reward being higher than expected, so things are going well) is used to push an agent to increased exploitation of current knowledge, whereas lower valence/reward (things are going worse than expected) pushes an agent to explore. This view is largely consistent with the {\em affect-as-cognitive-feedback} view~\cite{Huntsinger2014} where higher valence facilitates usage of existing mental constructs, whereas negative valence inhibits (and thus forces an agent to seek new solutions through exploration).  In general, expected and immediate emotions are hypothesized to be related to expected utility and modify action choices accordingly~\cite{Lerner2003}. In particular, emotions such as fear/hope/joy/distress have been related and computed  directly from value functions and rewards. For example, ``joy'' is defined as the likelihood of a change in expected value, while ``hope'' is defined as expected value (if positive)~\cite{Broekens2015}. Gomez and Insua use a decision theoretic definition of hope, fear, happiness and sadness, in much the same way~\cite{Insua2017}. These emotions are computed based on expected utility, and the elicited emotions' intensity is partially based on uncertainty If a threshold in uncertainty is crossed, then a short-circuit ``impulsive behaviour'' is triggered that overrides the rational one. Although the authors claim this is a ``System 1/System 2'' dual process approach~\cite{Kahneman11}, it is much more in line with the Simonesque ``interrupts'' approach~\cite{Simon67}. An adaptive system combining fuzzy logic with reinforcement learning is described in~\cite{ElNasr2000}. This model also uses application-dependent appraisal rules based on the OCC model~\cite{Ortony1988} to generate emotional states. However, they also use a set of ad-hoc rules to generate actions. A recent survey of the connections between emotional appraisals and elements in reinforcement learning is in~\cite{Moerland2017}.

%A number of modifications of the reward function based on appraisals are defined in~\cite{Sequeira2014}.

A generalisation of these approaches is intrinsically motivated reinforcement learning (IMRL), which is an evolutionarily plausible mechanism for allowing agents to learn a reward function that will lead them to be maximally fit~\cite{Singh04}.  The idea is to search the space of reward functions for one that maximizes the agent's fitness as defined by the extrinsic (usual, designer-provided) reward.  A distinction is made with normal ``reward shaping'' because of resource bounds or agent limitations: the optimized reward function in IMRL will take these bounds into account, finding the best possible reward function given the limitations of the agent.  Emotions have been proposed as defining the space of reward functions over which an IMRL method will search~\cite{Sequeira2014}. A set of emotional appraisal features are defined, the weights of which are exhaustively examined to determined the optimal setting for a given domain. The appraisal features used in~\cite{Sequeira2014} are novelty (defined in terms of how many times a state-action pair has been tried), goal relevance (defined as a heuristic estimate of the distance to a maximally valued state), controllability (with uncontrollability measured by the Bellman error), and valence (measured by the value function directly).  These appraisal features resolve to standard heuristic methods for guiding reinforcement learning, such as {\em optimism under uncertainty} (exploration bonus) or Bellman error (exploitation bonus), being now referred to as an emotional appraisal of {\em novelty} or {\em control}, respectively.   Even in the multi-agent case~\cite{Sequeira2011}, the appraisals are direct encodings of altruistic reward functions, leading to simple modifications to reward functions that cause rational agents to be more cooperative (as in~\cite{Nowak2006}).  The learned reward functions are domain-specific, and fail to generalise. Attempts to find an optimal ``universal'' reward function result in an agent that optimizes the extrinsic reward, with a small exploration bonus (encoded as a negative reward for controllability)  that decreases as the agent learns the environment~\cite{Sequeira2014}.  In \cite{Marinier08}, the SOAR cognitive architechture is augmented with a reinforcement learning agent that uses a synthesis of many emotional appraisals (termed a ``feeling'') as a reward signal. As in the above approaches, these emotional appraisals are direct encodings of decision theoretic primitives or heuristics (e.g. direction to goal).  Although Marinier and Laird~\cite{Marinier08} associate this approach with IMRL, it is not related as the agent does not attempt to match the external reward function but only uses its internal reward signal to learn from.

\subsection{Rationality and Superintelligence}
Athough it is increasingly clear that humans operate with something akin to a dual process model~\cite{ZhuThagard2002}, some (e.g. artificial intelligence practitioners) may argue that a connotative representation is unecessary for general intelligence, and that sufficient resources (relaxing the resource bound) will lead to fully denotative, decision-theoretically rational agents, or ``econs''. The higher precision allowed by a denotative representation seems to point the way to a superintelligence~\cite{Hofstadter1983}, and a decision-theoretically rational social system. We propose that artificial intelligence {\em requires} a dual process denotative/connotative model to exist as a general-purpose member of a socio-technical system. We present computational, evolutionary and social arguments here.

One argument is that an agent's ability to model the vast numbers of combinations of other agents becomes challenging unless a lower dimensional manifold is discovered that enables cooperation in groups. Consider a multi-agent system consisting of agents of $N$ different types % , $y_i; i\in \{1\ldots N\}$
who can behave in $N_B$ different ways. If each agent attempts to model all other agents in its group, including their first-level (direct) interactions with each other, the number of combinations would be factorial in the product $N\times N_B$. If  $N=150$~\cite{Dunbar1992} and $N_B$ is $500$ or so, then representation is intractable, and even  with only $N=N_B=10$, the number of combinations is astronomical.
%~\footnote{Indeed, with only $N=M=10$, the number of combinations is astronomical, approaching the number of atoms in the universe.}.

The second argument in support of a connotative state is offered by Turner~\cite{Turner2015} from an evolutionary perspective, who notes that early apes were forced into the forest canopy by other simians around 25 million years ago, and had to deal with a more complex, three dimensional space, making permanent groups more difficult and leading to a species with no permanent bonds and increased promiscuity. %where simply running quickly away from a predator doesn't always work.
%This, along with a host of other factors, forced these apes to develop (what are referred to here as) variational approximations to the (increased) true free energy presented.
This increased complexity entailed an increase in the true free energy (number of configurations the world can be in) that the apes had to model, and pressured the development of approximations. % to this increased free energy. % which we model here using variational Bayesian methods. 
When the descendents of these apes, the early hominids, were forced into the savannah in an era of climate change around 10 million years ago, the reduced complexity of a two dimensional world, combined with the need for stronger group cohesion (because of predators) pushed these approximations to other uses, fostered the development of early ``emotional'' languages, and allowed larger structures of humans to form, opening the door for collective activities like solving social dilemmas. From an evolutionary perspective, the perserverance of this emotional language is an indication of its usefulness in the context of human groups, and therefore we expect it to be useful in a group involving artificial agents as well.

Finally, a group of agents who are able to coordinate to solve social dilemmas will be more suited for survival than a group that does not. This coordination departs from the principles of decision-theoretic (individual) rationality, but can be enforced by a connotative representation that inextricably links agents through emotional signaling. This inextricable link provides a mechanism for a group of agents to jointly minimize their free energy in an uncertain world. %, and is thus a provable statement.

\section{The socio-psychological Control of Affect}
\label{sec:bact}
We present here a short introduction to ACT and \bact, and refer the reader to longer treatments in~\cite{HoeyBACT15,SchroederHoeyRogers2016}, covering relationships to other theories of emotion (e.g. appraisal).

\subsection{Affect Control Theory}

Affect Control Theory (ACT)~\cite{Heise2007,MacKinnon1994} %; MacKinnon and Heise 2010; Smith-Lovin and Heise 1988)
proposes a fundamental link between symbolic, denotative, representations of the social environment and the continuous, connotative, representations of the sentiments or feelings associated with those denotative representations. For example, when one perceives a person in a white coat in a hospital, a denotative impression is formed of this person that is represented with a symbol (“doctor”). This symbol has an associated ``fundamental'' sentiment in a three-dimensional affective ``EPA'' space of evaluation (E: good/bad), power (P: strong/weak) and activity (A: active/inactive). Doctors, for example, usually evoke feelings of goodness, strength, and modest activity.
%This space will be referred to as the ``EPA space'' in what follows.
EPA space has been found through decades of research to be a cross-culturally normative representation of meaning~\cite{Osgood1969}.

The link between denotative and connotative in ACT is empirically determined through population surveys using semantic differential scales. These measurements yield a set of samples from a population distribution in the sentiment space, which can then be parametrically estimated (e.g. as the mean and variance of a normal distribution), or non-parametrically represented (as a set of samples). Lists of such measurements are called ``dictionaries'' of mappings from labels to sentiment. In ACT, only the mean of this measurement is used to link denotative and connotative. Thus, a \identity{doctor} is represented connotatively as  \epa{2.7}{3.0}{0.2}.\footnote{For historical reasons, EPA measurements are scaled to lie between -4.3 and +4.3. All data in this paper is taken from a survey of 1742 people in the USA in 2015, see \url{https://research.franklin.uga.edu/act/}.}  Given a connotative (EPA) vector, a denotative label can be assigned in ACT using a simple nearest neighbour method (e.g. the closest label to \epa{-1.0}{2.0}{2.0} is \identity{politician} \epa{-0.9}{2.3}{1.5} - at a Euclidean distance of $0.35$). ACT proposes that events in the world, interpreted symbolically (denotatively), create re-assessments at the connotative level called {\em transient impressions} that are used to motivate agents towards behaviours that reduce the incoherence between in-context impressions and out-of-context sentiments. This motivation to socially conforming actions can be interpreted as an instance of Bourdieu's ``habitus''~\cite{Bourdieu1990}, as explored in more detail in~\cite{Ambrasat2016}.

Emotions in \act are defined precisely as the vector difference between fundamental (out-of-context) and transient (in-context) sentiments, and are a mechanism to help agents signal (in)coherence to each other (e.g. with facial expressions or paralinguistics). Importantly, these signals are not scalar indications of (in)coherence, but rather vector signals giving recipies for restorative behaviour and emotion regulation~\cite{Gross1998}. For example, if a doctor \identity{talks down to} \epa{-1.6}{-0.1}{0.3} another doctor, the object agent is made to feel less powerful (drops to $-0.1$) than expected, and will display exasperation or indigance. Upon receiving this signal, the acting agent may restore fundamental sentiments by \identity{making up with} the other.

%that deals with this problem using a clear and precise definition: emotion is a vector measure of perceived inconsistencies in sentiments caused by group behaviour. This vector measure is ``felt'' by humans and is often categorized {\em post hoc} using an emotion label (e.g. ``happy'' or ``angry''). Humans use their expressive modalities (e.g. their face) to signal the inconsistencies (emotions) to other members of the group.  These performative signals, interpreted by other group members, are used to {\em restore} the social order by resolving incorrectly inferred affective sentiments of group members that arise because of purposeful or exploratory behaviour, lack of information, or noisy communication channels.  If all group members are attempting to live at the equilibrium point, and some group member signals a perceived inconsistency, then this signal will be interpreted as a threat to the group equilibirum, and group members will be strongly motivated to resolve the inconsistency by re-interpreting identity or modifying behaviour.  This motivation arises evolutionarily because 

\subsection{\bact}
\bact~\cite{HoeyBACT15,SchroederHoeyRogers2016} generalises ACT by explicitly representing the distribution over sentiment in a two-level partially observable Markov decision process (POMDP). \bact models individual differences as variance in sentiments, and modulates the predictions of ACT due to the differences between denotative entities with low and high connotative variances~\cite{FreelandHoey2017}.  In the original formulation of the \bact model, the sentiment was directly observed in an interaction as a three-dimensional, continuous vector that gave a direct measurement of the sentiment of the behaviour being performed. That is, if a doctor was observed injecting someone with medicine, then \bact expected a direct observation of the mean EPA rating for that denotative behaviour, {\em inject someone with medicine}: \epa{0.9}{1.7}{-0.2}. \bact had a denotative state, but this only represented elements of an interaction outside of the social definition of identities, such as the state of a game being played. For example, this might be the positions of both agents' pieces on a chessboard, or current bids in a negotiation.

Here, we propose that the \bact model includes a denotative representation of identities and behaviours of other agents. In this case, these denotative elements are linked to the connotative state through a potential function that 
%This is an important generalization because it ensures intextricability by having both denotative and connotative present simultaneously.
%The somatic potential is defined in terms of an energy  function that
measures the incoherence (difference) between the current estimate of the denotative state (e.g. {\em doctor}) and the current estimate of the connotative state (a distribution in the affective EPA space).  We call this potential function the {\em somatic potential}
The somatic potential is defined in terms of an energy function that measures the incoherence (difference) between the current estimate of the denotative state (e.g. “doctor”) and the current estimate of the connotative state (a distribution in the affective EPA space).

For example, if the doctor performs some behaviour uncharacteristic of a doctor (e.g., {\em abuse} a patient), this doctor would seem less good (lower E) than the culturally accepted definition of a doctor. The incoherence generated between the out-of-context sentiment about doctors (high E) and the impressions created by the observed behaviour pushes the observing agent to a higher energy state. While behaviours can be selected (as in ACT) to reduce incoherence (and thus energy), the energy function can also be used to probabilistically rank likely identities that could be used for re-identification. Thus, if a doctor is observed {\em harassing} \epa{-3.0}{0.6}{1.6} a patient \epa{0.6}{-1.5}{-1.3}, agents would be motivated to act in such a way as to stop the behaviour, or would be forced to re-interpret the doctor as some other identity (the optimal in this case would be \epa{-4.3}{1.4}{1.7}, with a closest label of {\em rapist} at a distance of $0.44$).

In the following, we present a mathematical definition of the somatic potential and associated energy function. We start with the assumption that an agent must maintain an internal model of the world as a set of states making up a state space $\StateSpace$, which is factored into a denotative part, $\DenSpace$, that describes the ontological states of entities in the world, and a {\em connotative} part, $\ConSpace$, that describes the meanings of entities in the world. In ACT and \bact, the connotative space spans the three-dimensional vector space of EPA sentiments for identities (labels assigned to people) and behaviors (labels assigned to people's actions). For example, a particular agent's identity may be represented by some $X$, such that the word \identity{doctor} in the English language is represented as a particular value of that variable $X = x$. Similarly, $\Yb=y$ will represent the affective meaning (in EPA space) of those denotative entities. For example, a doctor might seem good and powerful \epa{2.7}{3.0}{0.23}. 

\bact has two sets of observations. One, $\omxb$, represents signals about the environment giving evidence for the denotative state. The other, $\omeb$,  represents emotional signals from other agents, and gives direct evidence for the connotative state. Information flows into the model from both connotative and denotative sides, and \bact computes posterior distributions that best merge the two in a Bayesian sense. Emotion signals are crucial for grounding the connotative state, as otherwise it could be arbitrarily transformed between agents and would be harder to learn.

\begin{figure}[htbp]
  % \begin{tabular}{cc}
  \begin{center}
    \resizebox{0.6\textwidth}{!}{\begin{tikzpicture}
[
randomvar/.style={circle, draw, scale=0.85, inner sep=1mm},
randomvarp/.style={circle, draw, scale=0.75, inner sep=1mm},
randomvarb/.style={circle, draw, scale=0.75, inner sep=1mm},
randomvarbp/.style={circle, draw, scale=0.65, inner sep=1mm},
actvar/.style={rectangle, draw, scale=0.85, inner sep=1mm},
obsvar/.style={rectangle, draw, scale=0.85, inner sep=1mm}
];

%%botleft (x,y) to top right (x',y')
%\draw (3,-3) to[grid with coordinates] (8,4);

%% UNPRIMED
%% connotative
\node (y) at (3, 3.5) [randomvar] {$\yb$};

%% denotative 
\node (x) at (3, 2) [randomvar] {$\xb$};

%% PRIMED
%% fundamentals
\node (yp) at (6, 3.5) [randomvarp] {$\yb'$};

%% denotative 
\node (xp) at (6, 2) [randomvarp] {$\xb'$};

%% ACTION
\node (a) at (4.5,2.3) [actvar] {$\ab$};
\node (fb) at (4.5,3.2) [actvar] {$\fub_b$};

%% OBSERVATIONS
\node (omxp) at (6,1) [obsvar] {$\omxb'$};
\node (omx) at (3,1) [obsvar] {$\omxb\phantom{'}$};
\node (omep) at (6,4.5) [obsvar] {$\omeb'$};
\node (ome) at (3,4.5) [obsvar] {$\omeb\phantom{'}$};

%% LINKS
%% probabilistic temporal
\draw [->] (y) -- (yp) [very thick];

\draw [->] (x) -- (xp) [very thick];

%% action changes denotative behaviour only
\draw [->] (a) -- (xp) [very thick];
\draw [->] (fb) -- (yp) [very thick];

%% state->observation links
\draw [->] (x) -- (omx) [very thick];
\draw [->] (xp) -- (omxp) [very thick];
\draw [->] (y) -- (ome) [very thick];
\draw [->] (yp) -- (omep) [very thick];

%% weight links

%% somatic transform
\draw (yp) --  (xp) [black!40!green,very thick];
\draw (y) --  (x) [black!40!green,very thick];
\draw (a) --  (fb) [black!40!green,very thick];

\end{tikzpicture}}
  \end{center}
    %&  \resizebox{0.3\textwidth}{!}{\input{../bact/stpic-full.tex}}\\
    %(a) & (b)\\
    %\end{tabular}
  \caption{\label{fig:stpic} Belief network for \bact at a high level of abstraction showing denotative $\xb$ and connotative $\yb$, observations $\omxb$, emotions $\omeb$, and actions both denotative $\ab$, and connotative $\fub$. Two somatic transforms link state and action, respectively, but can be considered the same. Primed variables are post-event, and the network is dynamically unrolled through time.}
  \end{figure}
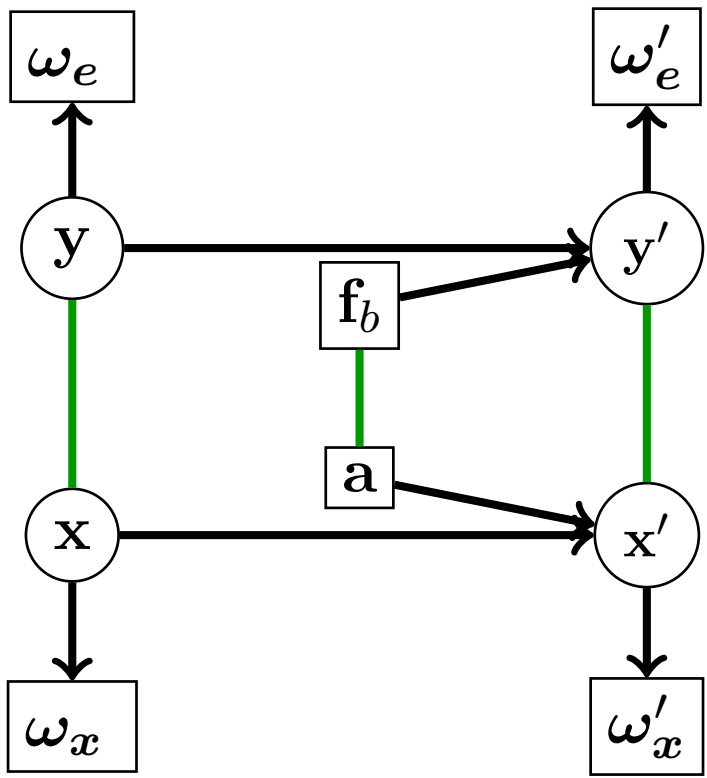

Finally, \bact has two sets of actions representing denotative action, $\ab$, and connotative meanings of those actions, $\fub_b$.  Figure~\ref{fig:stpic} shows a Bayesian network representation of \bact.

\subsection{Somatic Transform}
A core element in the \bact POMDP is that every denotative element $\Xb$ (including agent actions, $\ab$) has an associated connotative element $\Yb$ ($\fub_b$ for actions). The important part is the connection between denotative and connotative elements, which we will write using a function called the somatic potential: $\Gub(\Xb,\Yb)$. The somatic potential specifies the shared cultural connotative interpretation of the denotative state $\Xb$. That is, for some $\xb$, $\Gub(\xb,\Yb)$ is a function over the connotative space, $\ConSpace$, representing the shared sentiments for denotative state $\xb$. For example, $\Gub(X_a=doctor,\Yb)$ could be a normal distribution given by summary statistics (mean, covariance) measured in a population survey. Using a somatic potential, agents can ask questions such as ``what do I feel about object $\xb$?'' (e.g. Q: ``how do you feel about doctors?''; A: ``I see doctors as quite good, quite powerful, and a bit active, but I'm somewhat uncertain about this''). 

The somatic potential can also yield, for some $\yb$, a function over the denotative space, $\DenSpace$, $\Gub(\Xb,\yb)$. For some discrete set of $N$ denotative entities, $\Xb=\{\xb_1,\xb_2...\xb_N\}$, this could be a multinomial distribution (a set of numbers $p_1,p_2,...p_N$ such that $0\leq p_i \leq 1\:\:\forall i$ and $\sum_i p_i = 1$) indicating that entity $\xb_i$ is likely with probability $p_i$. The agent can therefore also ask questions such as ``what sort of $\xb$ feels like $\yb$?'' The reader can demonstrate that this is the ``fast'' or ``System 1'' thinking at work by playing the following game. First, pick values for E,P and A  and then imagine words with those EPA values.  Notice how quickly these very different things come to mind. (e.g. Q:''what sort of thing is very good, a bit weak, and very active?''; A: “dogs, smiley faces, kids”).
%(e.g. Q:``what sort of thing is very good, a bit powerful and very active?''; A: ``dogs, simley faces, kids''). 

To give the somatic potential a more precise definition, we may model it as a joint probability distribution over $\Xb$ and $\Yb$, written as a Boltzmann distribution~\footnote{The Boltzmann distribution is often used in statistical physics to describe the joint probability of different states (e.g. of a gas or spin system). It is usually represented as $P(x) = \frac{1}{Z} e^{-(E(x))/kT}$ where $E(x)$ is the energy of configuration $x$, $T$ is the temperature and $Z$ is a normalizing factor called the ``partition function''. Estimating the partition function (and thus knowing the actual probability of an event $x$) is challenging because it is a sum of energy terms across all possible states (many of which may not even be known by the agent). The partition function is equal to the free energy of the system scaled by the temperature, $T$. The free energy is a measure of how complex the environment or system is. We use the Boltzmann distribution here for convenience. It could be replaced with other distributions.}
of the energy function defined earlier, which we write as $E(x,y)$, denoting the energy or incoherence between the denotative and connotative states :
\begin{equation}
  G(x,y) = c e^{-E(x,y)} = c e^{-(y-M(x))^2/\gamma^2}
  \label{eqn:boltzmann}
\end{equation}
Where $M(X=x)$ is a function in $Y$ giving the connotative distribution for that particular $x$ and $c$ is a normalizing constant (the inverse of the partition function). M may be simply the mean of the population survey for the concept $X=x$, and $\gamma$ a constant. Thus, the more coherent $x$ is with $y$, the smaller the energy $E(x,y)=(y-M(x))^2/\gamma^2$, and the more likely the state. The parameter $\gamma$ models one aspect of the (emotional) predictability of the environment. As the environment's diversity increases (say with the addition of some heterogeneous other agents), $\gamma$ naturally increases as the ascription of sentiment to denotative elements in the world is less well defined. Such a world becomes less predictable and less valid. Choosing a value for $\gamma$ is a learning choice to be made by an agent. Although we know that $\gamma$ is related to the variance in sentiments in the population, it does not need to be exactly the same. In the Georgia dataset, the variance in power for \identity{doctor} is $1.4$, while that for \identity{nurse} is $2.5$. However, the value of $\gamma$ will also be a function of the agent's social network, as the agent may operate in a clique or cluster of locally more homogeneous sentiments.

For ease of exposition, here we will assume that the connotative state is one dimensional, and we focus exclusively on the somatic transform. The dynamics in the full \bact POMDP are more fully explored elsewhere. Here, we suppose that a prior marginal over $X$, $P(x)$, represents the agent's current estimate of the denotative nature of the interaction.\footnote{Here we show one possible usage of the model, assuming prior marginal distributions and attempting to find posterior marginals. We assume the joint probability over $X$ and $Y$ may be modeled as a product of the two marginals. Maintaining this independence is important for the stability of the model, allowing for two, separate, but related systems. It also allows for the two systems to operate at different time scales (with the connotative system able to make faster predictions than the denotative system), and is more in line with neuro-biological findings about different brain systems devoted to the two systems. Note that the time scale difference is actually a difference in noise levels, with the denotative system making worse predictions in less valid environments, so the connotative system ``takes over'' while it ``waits''. It is therefore the rate of change of uncertainty that matters, and is externalized as a ``fast'' and a ``slow'' system.} It can be a belief about various properties of objects being manipulated as part of the ongoing interaction, for example. Thus, an agent may observe a female in a white coat in a hospital setting, and infer a distribution over the possible identities of \identity{nurse} or \identity{doctor}. In this case, an agent with a gender stereotype may form a denotative impression which puts more mass on \identity{nurse} than on \identity{doctor}\footnote{Research shows patients have fairly strong opinions about what doctors and others should be wearing~\cite{Petrillie021239}}. Such denotatively constructed impressions can be made with constraint networks (e.g.~\cite{FreemanAmbady2011,JosephMorgan2019}) or other relational models such as relational databases. In our model, the result of the constraint satisfaction convergence or stabilization is $P(x)$, and would be represented in this simple case with one number $p$ giving the probability that this entity (person) is a \identity{nurse}, for a distribution over [\identity{nurse}, \identity{doctor}] of~\footnote{Clearly there may be more than two identities under scrutiny at a time (e.g. head nurse, orderly, medical student, etc). We proceed here with two without loss of generality.}

\[P(x) = P([nurse,doctor])=[p,1-p]\]

Suppose further that a prior marginal distribution over $Y$, $P(y)$, represents an agent's current estimate of the affective nature of the interaction. It can be a belief about identity sentiments of participants, recent or forthcoming behavior sentiments, or sentiments about settings or other physical objects.  Suppose the agent had observed the same female in the white coat performing a gesture implying she has power, such as ordering someone to do something. In this case, the agent's prior over the white-coated female, $Y$, would be shifted towards more powerful values.

\begin{figure}[htbp]
  \begin{center}
    \begin{tabular}{cc}
\resizebox{0.19\textwidth}{!}{
\begin{tikzpicture}
[
randomvar/.style={circle, draw, scale=0.85, inner sep=1mm},
];

%%botleft (x,y) to top right (x',y')
%\draw (3,-3) to[grid with coordinates] (8,4);

%% UNPRIMED
%% connotative
\node (y) at (3, 3) [randomvar] {$\yb$};

%% denotative 
\node (x) at (3, 2) [randomvar] {$\xb$};

%% LINKS
%% somatic transform
\draw (y) --  (x) [black!40!green,very thick];

\end{tikzpicture}
}
  &

\resizebox{0.4\textwidth}{!}{
\begin{tikzpicture}
[
randomvar/.style={circle, draw, scale=0.85, inner sep=1mm},
randomvars/.style={circle, draw, scale=0.85, inner sep=1mm},
];

%%botleft (x,y) to top right (x',y')
%\draw (3,-3) to[grid with coordinates] (8,4);

%% UNPRIMED
%% connotative
\node (y) at (3, 3) [randomvar] {$\yb$};

%% denotative 
\node (x) at (3, 2) [randomvar] {$\xb$};

%% G
\node (G) at (4,2.5) [randomvars] {$G$};
%% LINKS
%% somatic transform
\draw [->] (y) --   (G) [black!40!green,very thick];
\draw [->] (x) --   (G) [black!40!green,very thick];

\end{tikzpicture}
}
    \\
    (a) & (b) 

\end{tabular}
    \end{center}
  \caption{\label{fig:bn-sat} Somatic Potential as a belief network (a) undirected, (b) equivalent directed graph introducing observed variable $G=true$.}
  \end{figure}
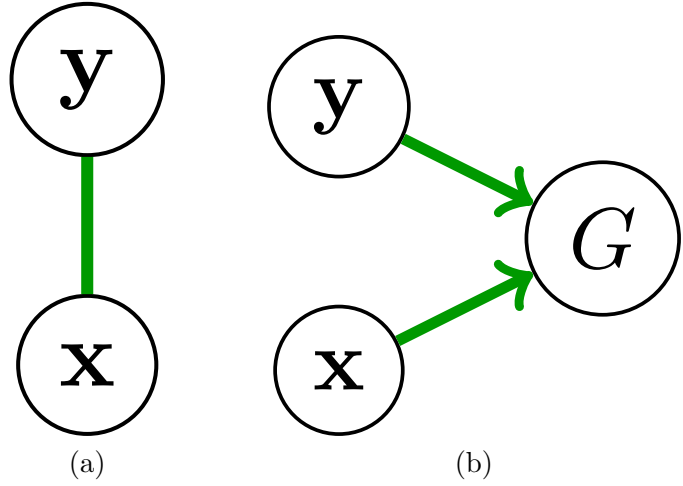

Using a joint prior that is factored as $P(\Xb,\Yb)=P(\Xb)P(\Yb)$, we seek a posterior distribution $P'(\Xb,\Yb) = P(\Xb',\Yb')$, which combines priors with the constraint imposed by the somatic potential (Equation~\ref{eqn:boltzmann}). Figure~\ref{fig:bn-sat}(a) shows a graphical representation of the somatic potential as a graphical model connecting two variables $\Xb$ and $\Yb$ with an undirected link which represents the potential $P(\Xb,\Yb)$. Note that we write both distributions and density functions as $P(\cdot)$, as the type of function and operations used is defined by the variable in context $\Xb$ or $\Yb$.

In fact, we are seeking the posterior {\em given our knowledge of Equation~\ref{eqn:boltzmann}}. We postulate a variable $G$ representing our knowledge that a somatic potential such as Equation~\ref{eqn:boltzmann} connects denotative ($\Xb$) and connotative ($\Yb$) spaces. We therefore know that this variable has value ``true'', which we write as $G=g$ or simply $g$, with the interpretation that $g$ {\em iff} the somatic potential exists between $\Xb$ and $\Yb$. We then are actually considering the joint distribution (after primes are removed) as $P(\Xb,\Yb|G=true)=P(\Xb,\Yb|g)$, shown in Figure~\ref{fig:bn-sat}(b). This distribution can be factored as
\begin{equation}
  P(\Xb,\Yb|g) \propto P(g|\Xb,\Yb)P(\Xb,\Yb) = \Gub(\Xb,\Yb)P(\Xb)P(\Yb)
\end{equation}
and we see that the joint distribution is in fact a product of the somatic potential and the prior distribution (which may be factored into two prior distributions over $\Xb$ and $\Yb$).
%This transformation will be used in Section~\ref{sec:bact:dynamics}.
Once $g$ is added, however, the undirected belief network is transformed into a directed network, as shown in Figure~\ref{fig:bn-sat}(b). This also helps practically for integration into the POMDP framework.

We can further integrate denotative and connotative evidence for each $X$ and $Y$, respectively with the addition of observables $\Omxb$ and $\Ome$, generated by $\Xb$ and $\Yb$, respectively, as shown in Figure~\ref{fig:bn-sat}(a). Thus, the model is agnostic to the {\em type} of sensor information given. 

With a somatic potential defined to be the Boltzmann distribution as above, and factored priors $P(x)$ and $P(y)$, an estimate of the marginal posterior over $Y$ ($P'(y)$, the feelings evoked after an event) can be computed as:

\begin{equation}
  P'(y)=\sum_xP'(x,y) = \sum_x P(y)P(x)G(x,y) \propto P(y)\sum_x P(x)e^{-(y-M(x))^2/\gamma^2}
  \label{eqn:ppy}
  \end{equation}

where the $\propto$ shows the quantities are proportional (equal up to a constant multiplier, labeled $c$ in Equation~\ref{eqn:boltzmann}). Thus, the posterior distribution over $Y$ is the expectation of the somatic potential with respect to the prior distribution over $X$, $P(x)$, multiplied by the prior distribution over $Y$.

Similarly, the posterior distribution over $X$, $P'(x)$, is given by 

\begin{equation}
  P'(x) = \int_y P'(x,y)dy = \int_y P(y)P(x)G(X,y) dy \propto P(x)\int_y P(y)e^{-(y-M(x))^2/\gamma^2} dy
  \label{eqn:ppx}
\end{equation}

The posterior distribution over $X$ is the expectation of the somatic potential with respect to the prior distribution over $Y$, $P(y)$, multiplied by the prior distribution over $X$. The ability to transform between connotative and denotative states (in either direction) will be referred to as a {\em  somatic transform}. 

Equations~\ref{eqn:ppy} and~\ref{eqn:ppx} can be analytically computed in the case of Gaussian priors, which we pursue below. In practice, the somatic transform can be problematic because it generates a posterior over $\Yb$ which is no longer a single Gaussian, but a sum of Gaussians. Projecting this very far into the future may lead to an explosion of modes. However, modes can be combined or rejected by action selection as well, meaning that for each sum of Gaussians generated, one can be selected through action. In the doctor example, an sum of two Gaussians results after a single iteration but the act of deference performed can resolve much of this uncertainty by committing to one hypothesis or the other.  %% possibly: show this? 

\subsection{Somatic Transform Examples}
\begin{figure}
  \begin{center}
    \includegraphics[width=0.8\textwidth]{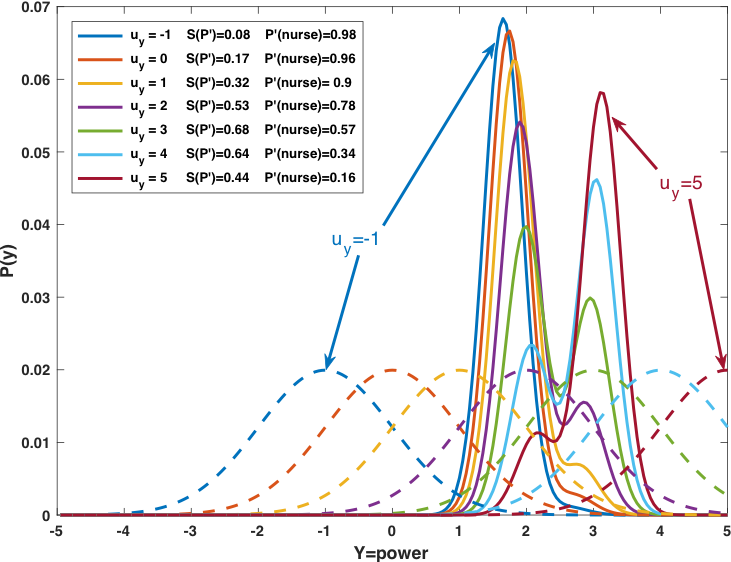}
    \end{center}
    \caption{\label{fig:st-sim-uy}  Effects of the Somatic Transform on the marginals over $X$ and $Y$. Gaussian priors over $Y$ are shown as dashed lines for different values of $\mu_y$. The prior over $X$ is $P(X=''nurse'')=0.7$. The posterior over $Y$ is shown as solid lines, while the posterior over $X$ is shown in the legend, with $S(P')$ denoting the entropy of $P'(X)$ and $P'(nurse)$ denoting $P'(X=nurse)$. As the prior shifts to more positive values in Y, the posterior in $Y$ shifts to be more in line with the power sentiment about \identity{doctor}, rather than \identity{nurse}. Further, the posterior in $X$ also favors \identity{doctor} (that is, $P'(nurse)\rightarrow 0.0$).
      }
\end{figure}

\begin{figure}
  \begin{center}
    \includegraphics[width=0.8\textwidth]{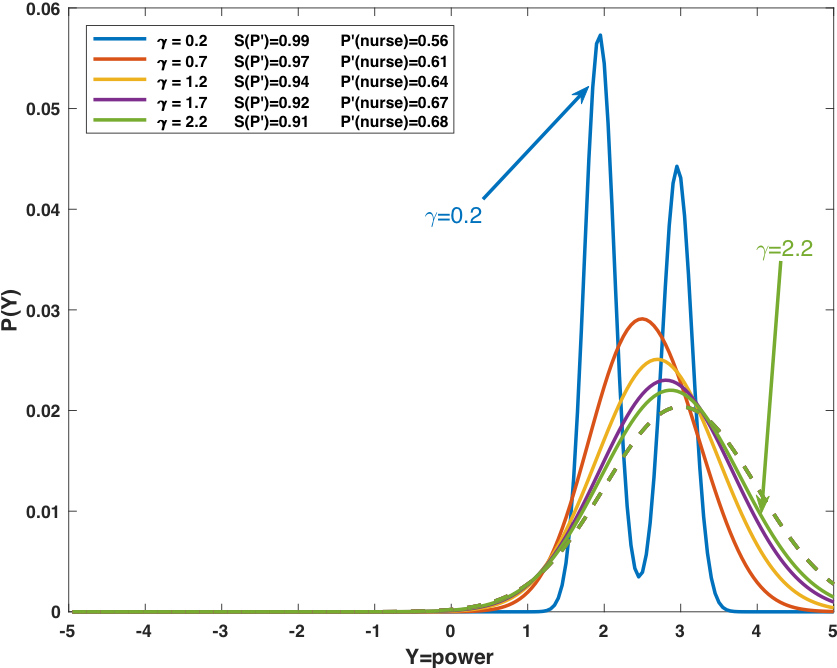}
    \end{center}
  \caption{\label{fig:st-sim-gamma}  Posterior over $Y$ with varying $\gamma$. As $\gamma$ decreases, the posterior over $Y$ is more focussed on the priors over $x$. $S(P')$ denotes the entropy of $P'(X)$. $P'(nurse)$ denotes the posterior probability of $X$ being \identity{nurse}: $P'(X=nurse)$.}
\end{figure}

Figure~\ref{fig:st-sim-uy} shows an example usage of this transform for a simple case.  In it, we consider a simple prior over $y$, $P(y)$, as a Gaussian distribution with a variance $\sigma_y=2.0$ and a mean $u_y$ which is varied to see the effect of a changing connotative prior. These distributions are shown as dashed lines in Figure~\ref{fig:st-sim-uy} (with means ranging from $-1$ to $5$). A prior over $x$, $P(x)$, represents only two identities \identity{nurse} and \identity{doctor} with probabilities $0.7$ (nurse) and $0.3$ (doctor). A-priori, the agent believes it is more likely this person is a nurse (possibly due to a complex constraint satisfaction network~\cite{FreemanAmbady2011}). The somatic transform is implemented using normal distributions as the values of $M$ mapping a label in $X$ to a mean and variance in $Y$ given by the Georgia 2015 survey data. The identities of \identity{nurse} and \identity{doctor} have power (P) values of $1.9$ and $2.95$, respectively. We only consider a single dimension here for ease of exposition, but the results carry over to $3$ or more dimensions. In Figure~\ref{fig:st-sim-uy}, we use $\gamma=0.3$, but investigate how $\gamma$ effects the results in Figure~\ref{fig:st-sim-gamma}. 

Figure~\ref{fig:st-sim-uy} shows how the posteriors evolve as $\mu_y$ is changed. The entropy in the posterior distribution over $x$, $P'(x)$, is shown as $S(P')$ in the legend,\footnote{Entropy is a measure used in statistical physics, and it describes the level of homogeneity of a distribution. In information theoretic terms, entropy measures the amount of information in a system that exists in a set of states $x$ according to the probability distribution $P(x)$. Entropy is typically written as $S(P) = -\sum_x P(x)\log P(x)$. A low entropy system has a joint distribution over states $P(x)$ which is not dispersed evenly across all $x$, and so is easier to predict. Kahneman and Klein~\cite{KahnemanKlein2009} refer to this as the ``validity'' of the environment.} along with the value of $P'(X=nurse)$ as $P'(nurse)$. The posteriors over $y$, $P'(y)$ are shown as solid lines. First, we can see that as the prior over $y$ approaches the prior over $x$ (with an expected Power sentiment value of $0.7\times 1.9+0.3\times 2.95 = 2.2$), the posterior becomes a bimodal distribution with about $70\%$ of its mass nearer to the \identity{nurse} identity at $1.9$. Further, as the priors more closely agree (that is $\mu_y$ approaches $2.2$), the entropy of the resulting distribution over $x$ increases, so the information obtained by combining them is smaller. For largely different values of the mean of $P(y)$ (e.g. $\mu_y=-1.0$ or $\mu_y=5.0$), the resulting entropy of $P'(x)$  is small, and more information was gained by the denotative system from the connotative system. The resulting distributions over $x$ are also shown, demonstrating a clear shift from \identity{nurse} ($P'(x)\rightarrow 1.0$) to \identity{doctor} ($P'(x)\rightarrow 0.0$) as the prior information about the sentiment observed shifts towards the positive (the person demonstrates a behaviour with more power).

Figure~\ref{fig:st-sim-gamma} shows how the same curves evolve according to a changing value of $\gamma$, with a fixed $\mu_y=3.0$, $P(x)=0.7$. As the world becomes less predictable ($\gamma$ increases), the connotative system increasingly steps in to help out. As Figure~\ref{fig:st-sim-gamma} shows, this happens naturally in the somatic transform. When $\gamma$ is large ($>2.0$ in this example) there is not as strong an effect between $X$ and $Y$, and so both follow their prior distributions more closely. For small $\gamma$, the sentiment follows the prior over $X$ much more closely, becoming more centered around the known mean values of power for the identities of \identity{nurse} and \identity{doctor} of $1.89$ and $2.95$. An agent using smaller values of $\gamma$ are therefore more likely to attribute fixed sentiments to individuals, requiring a lessening of heterogeneity. Such agents believe that other agents should behave in less flexible ways, but trust in a more valid environment to provide them with the ability to predict denotatively. At the other extreme of large $\gamma$, agents believe other agents are more flexible, but more consistent connotatively.

\begin{figure}
  \begin{center}
    \includegraphics[width=0.8\textwidth]{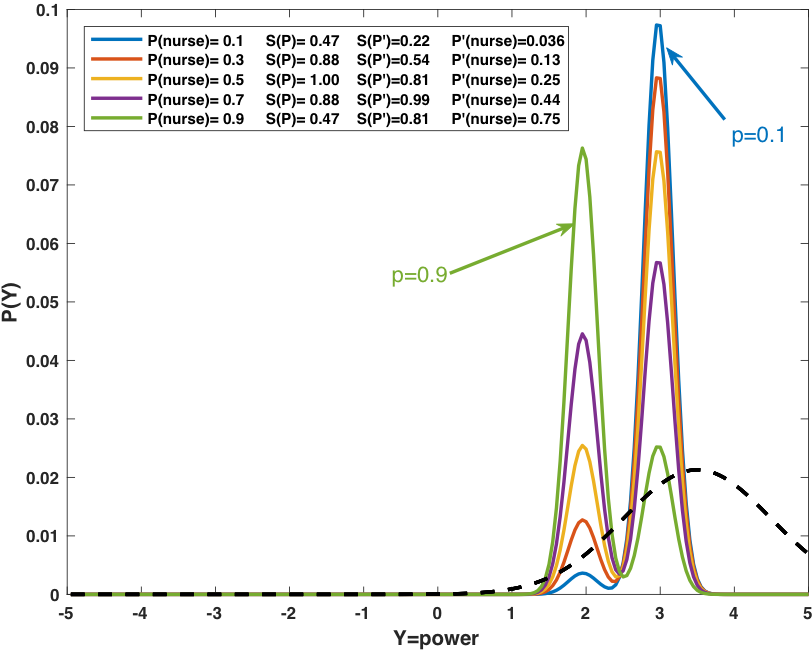}
    \end{center}
  \caption{\label{fig:st-sim-px} The top (blue) and bottom (green) lines in the legend show a prior state with a less dispersed $P(X=''nurse'')=p$, with $p=0.1$ and $p=0.9$, respectively, yielding a posterior for both $X$ and $Y$ that is more in line with the original denotative prior $P(x)$. The orange line ($p=0.5$) shows how the posterior is biased towards the prior in $y$ (possibly based on stereotypes). The prior in $y$ is shown as a black dashed line (same for all values of $p$). $S(P)$ and $S(P')$ denote the prior and posterior entropy of $P(X)$, and $P(nurse)$ and $P'(nurse)$ denote the prior and posterior probability of $X$ being \identity{nurse}.}
\end{figure}

The somatic transform naturally shows a trade-off between the uncertainty in $X$ and $Y$.  Figure~\ref{fig:st-sim-px} is an example of this trade-off showing how the posterior over $X$ and $Y$ changes as a function of $p=P(X=''nurse'')$. In this simulation $\sigma_y=3.5$  and $\gamma=0.2$. As the environment becomes less valid (less predictable or more uncertain, so $P(x)$ is more dispersed or has higher entropy), then $P'(y)$ and  $P'(x)$ will be more heavily influenced by the prior in $y$. Thus, when $p=0.5$, $P'(x)<0.5$. Agents in less valid (less predictable) environments will put more weight on the connotative system: they will make inferences and choose actions that are more in line with connotative (socio-cultural) expectations. In more valid environments, $P(x)$ has lower entropy and thus dominates the posterior, leading to a posterior that is more heavily influenced by the denotative state. Agents in more valid environments will thus act more in line with denotative concepts and predictive dynamics, and so will be information seekers and utilizers. In a social dilemma, for example, one would expect the agents in less valid environments to cooperate (act according to social prescriptions), while agents in more valid environments will defect (act decision theoretically rationally). This is in line with experiments showing how humans tend to act more pro-socially (cooperate in a public goods game) in ambiguous situations (ones in which risk is hard to evaluate, see Vives and FeldmanHall, 2018). In this simulation, risk is $p$: if $[p,1-p]$ is lower entropy, then risk is more well defined, and so ambiguity (the uncertainty in risk) is lower.

\subsection{Relationship of \bact and ACT}
Note that in ACT, the somatic transform has a zero temperature parameter $\gamma=0$. Further, either $P(x)$ is a point estimate ($x_o$), and $P(y)$ is a constant (no prior, set to $1$), or $P(y)$ is a point estimate ($y_o$) and $P(x)$ is a constant. Writing a point estimate as a delta function, $\delta(x - x_o)$, where  $\delta(x - x_o )=1$ if $x =x_o$   and $0$ otherwise, then in the first case, Equation~\ref{eqn:ppy} is:
\begin{equation}
  P'(y)= \sum_x\delta(x - x_o )G(x,y) \propto e^{-(y-M(x_o ))^2/\gamma^2 }=\delta(y - M(x_o))                   
\end{equation}

And in the second case, Equation~\ref{eqn:ppx} is:

\begin{equation}
  P'(x)= \int_y\delta(y - y_o)G(x,y)\propto e^{-(y_o-M(x))^2/\gamma^2}=\delta(y_o - M(x))
  \label{eqn:ppxn}
  \end{equation}

which simply says that $x_o$ and $y_o$ are related through the function $M$ directly (e.g. $M$ is a dictionary linking each $x_o$ with a $y_o$). Technically in equation~\ref{eqn:ppxn} this assumes a dictionary with an entry for every possible $y_o$, clearly an impossibility. Finding the nearest neighbor, as described above, is one possible way to circumvent this.

As mentioned previously, the somatic transform is the key difference between our presentation of the model here and the presentation of it in the original exposition of the BayesACT model~\cite{HoeyBACT15,SchroederHoeyRogers2016}. In the original presentation, we assumed that behaviors were both perceived and generated in the connotative/affective-space, leaving the translation to/from denotative spaces to some other perceptual or motor system. The somatic transform mathematically defines this translation and integrates it directly and deeply into the model itself. Rather than perceiving a behavior as a vector in a 3D affective space, an agent perceives and cognitively interprets denotative aspects of the situation including behavior, then uses these denotative aspects as evidence in support of its connotative/affective predictions, computing the level of support using a somatic transform. Simultaneously, connotative predictions are mapped to future denotative states and actions, providing heuristic guidance to an agent. This aspect of the original presentation of the BayesACT model is thus revealed as a simplifying assumption that has been replaced with the more general somatic transform.

%%CHECK- needs a rewrite.
The somatic transform thus captures the inextricability of $X$ and $Y$ (denotative and connotative). The transform fits into the \bact partially observable Markov decision process (POMDP), and therefore becomes an integral part of that model. The primary additional elements in BayesACT are (1) the temporal (dynamic) nature of both $X$ and $Y$; (2) the ability to construct denotative plans of action; and (3) the observation of X in the world through some set of sensors. While (1) and (2) are important and define how the agent will plan and act, it is not the subject of this article. The observation function, however, is a more statically defined element, and is considered to lie at the reactive level. It may contain further sub-programs that take care of very rapid responses (to immediate threats, for example). Thus, the reactive level of Ortony, Norman and Revelle~\cite{Ortony2005} could be captured by a lower level Bayesian model that also includes a policy of action. This would be summed up at the level of BayesACT using some probability distribution over possible observations, $O$, given the denotative interpretation, $X$, as $P(O|X)$. Further, the policies of action are represented in the dynamic process over $X$, with more likely futures being exactly the ones predicted by the computed policies. %We return to this in our simulations of conformity experiments. 

\section{Exploratory Examples}
The simple model above with a varying $P(x)$ provides an explanation for three related behavioral effects noted in the literature. First, Van-den-Bos~\cite{Bos2001} showed how thoughts of uncertainty about the self can lead to more pro-social behaviour. We show in Section~\ref{sec:bos} how this is a tradeoff from denotative to connotative in \bact. Second, in a classic experiment, Festinger gave participants (teenage girls) one of two prizes of equal value to them (audio records of unknown pop stars). The participants subsequently raised their evaluations of the prize they obtained. In general, a person is given one of two items that she values about the same, then she will value the item she is given more highly in order to reduce the cognitive dissonance created by the fact that she did not get the other item. This is also an example of the “confirmation bias” in behavioral economics, in which people seek explanations that confirm their prior beliefs. In Section~\ref{sec:cogdis}, we show that, according to the somatic transform and \bact, such re-interpretation of value is simply the process of attempting to unify connotative representations of the self (e.g. ``I am a good person'') with denotative representations of uncertain events (e.g., ``I think my prize is worse than hers'').

The third experiment we consider in Section~\ref{sec:conform} is the equally well known Asch conformity tests~\cite{Asch1951}, in which agents are led to denotative choices that are clearly wrong by a bias introduced by other agents making incorrect decisions. In \bact this is modeled again as a denotative/connotative tradeoff. As more other agents agree with the wrong choice, the denotative solution breaks down, leading agents to fall back on connotative meanings and conformity (in-groupness).

\subsection{Uncertainty and fairness}
\label{sec:bos}
Van~den~Bos~\cite{Bos2001} carried out an experiment in which the fairness or unfairness of a situation was evaluated affectively (positive vs. negative) in two conditions: one with induced (primed) feelings of uncertainty enhanced by asking about the emotions felt during uncertain episodes. Two effects are shown. First, the more (perceived) fair condition (where participants got to voice their opinion about a distribution of payoffs) elicited more positive emotions than the non-fair condition (no chance to voice). Second, the effect was {\em enhanced} by the increased uncertain feelings. In \bact this can be accounted for by noting that as the emotions associated with uncertainty about the self are evoked, so is the uncertainty in the denotative identity. %(due to AGP-ARP connections~\cite{Smith2019} - %%CHECK EXPAND THIS HERE).
As uncertainty about denotative identity is increased, the participant (a student) will be more reliant on the connotative system. %which says that student prefers the fair solution, but is not all that strongly affected by the unfair one.

Consider a purely denotative solution. In this case, the emotions elicited will not play a role, and a student will think that voicing an opinion may change the payoffs in their favour, and so will prefer that option. However, the other option of letting the experimenter decide is not a big deal, so they would rank the voicing condition as better (as they may assign a small probability to their voicing having an impact), but only marginally so. On the other hand, a purely connotative system will ignore the main identity of student and focus on a transient created by the student identity modified by an emotion of anxiety. Such an identity (``anxious student'') will be much more likely to prefer the fair option because it is the one that goes the furthest in restoring connotative meanings to the system.

We show the simplest possible example, using only ACT equations in order to motivate the problem. We show how \bact would modify things at the end of the section.  In~\cite{Bos2001}, there are $2\times 2$ conditions, with half the participants having a ``voice'' and half not, and half of them having uncertainty made salient, half not.
%If we plot the evaluation dimension (as measured by the PANAS) as a function of the voice/no voice condition, we get one line for salient, and one for non-salient as shown here:

\begin{figure}
  \begin{tabular}{cc}
    \resizebox{0.5\textwidth}{!}{\begin{tikzpicture}

  %axes y axis
  \draw [->] (1,0) -- (1,5) [thick];
  %% x axis
\draw [->] (1,0) -- (10,0) [thick];

%lines for distance from ``sad''
%\draw [-] (2.5,1.07) -- (7.5,3.27) [very thick];
%\draw [-] (2.5,1.33) -- (7.5,1.64) [dashed,very thick];
%% lines for average of distances from sad and disappointed
%% sad: -1.88
%% disappointed -1.71
%% we compute the e value for the four conditions
%% non-salient: its the ``E'' value of the characteristic emotion of ``student'' (1.5,0.66,0.26) ``warm'' ``easygoing'' in the no-voice condition, and the ``E'' value of the emotion felt when ``student'' compromises with ``student'' in the voice condition.
%% salient: same as above but for ``anxious student'', with a characteristic emotion of (-0.84,-0.14,0.11) ``envious'' in the no-voice condition or 2.31,1.24,0.04 in the ``voice'' condition.
%% ((((sad-e)+4.3)*6/8.6+1)+((disapointed-e)+4.3)*6/8.6+1)/2
\draw [-] (2.5,1.13) -- (7.5,3.33) [very thick];
\draw [-] (2.5,1.39) -- (7.5,1.70) [dashed,very thick];

%ticks and values
\draw [-] (0.5, 0) -- (1,0) [very thick];
\node at (0, 1) [scale=1, text centered,xshift=-1cm] {1};
\draw [-] (0.5, 1) -- (1,1) [very thick];
\node at (0, 2) [scale=1, text centered,xshift=-1cm] {2};
\draw [-] (0.5, 2) -- (1,2) [very thick];
\node at (0, 3) [scale=1, text centered,xshift=-1cm] {3};
\draw [-] (0.5, 3) -- (1,3) [very thick];
\node at (0, 4) [scale=1, text centered,xshift=-1cm] {4};
\draw [-] (0.5, 4) -- (1,4) [very thick];

%action labels
\node at (7.5, -0.5) [scale=1, text centered] {no voice};
\node at (2.5, -0.5) [scale=1, text centered] {voice};
\node at (7.5, 1.7) [scale=1, text centered,xshift=1cm] {non-salient};
\node at (7.5, 3.3) [scale=1, text centered,xshift=1cm] {salient};

\end{tikzpicture}}
    &
    \resizebox{0.5\textwidth}{!}{\begin{tikzpicture}

  %axes y axis
  \draw [->] (1,0) -- (1,5) [thick];
  %% x axis
\draw [->] (1,0) -- (10,0) [thick];

%liens
\draw [-] (2.5,1.8) -- (7.5,3.8) [very thick];
\draw [-] (2.5,2.2) -- (7.5,2.8) [dashed,very thick];

%ticks and values
\draw [-] (0.5, 0) -- (1,0) [very thick];
\node at (0, 1) [scale=1, text centered,xshift=-1cm] {1};
\draw [-] (0.5, 1) -- (1,1) [very thick];
\node at (0, 2) [scale=1, text centered,xshift=-1cm] {2};
\draw [-] (0.5, 2) -- (1,2) [very thick];
\node at (0, 3) [scale=1, text centered,xshift=-1cm] {3};
\draw [-] (0.5, 3) -- (1,3) [very thick];
\node at (0, 4) [scale=1, text centered,xshift=-1cm] {4};
\draw [-] (0.5, 4) -- (1,4) [very thick];

%action labels
\node at (7.5, -0.5) [scale=1, text centered] {no voice};
\node at (2.5, -0.5) [scale=1, text centered] {voice};
\node at (7.5, 2.8) [scale=1, text centered,xshift=1cm] {non-salient};
\node at (7.5, 3.8) [scale=1, text centered,xshift=1cm] {salient};

\end{tikzpicture}}
    \end{tabular}
  \caption{\label{fig:bos}(a) ACT simulations of conditions, showing the scaled average of the distance from the emotion felt in thecondition with the evaluation of the emotion of \identity{sad} and \identity{disappointed}. Scaling to the range 1-7 is done to match scales with that of~\cite{Bos2001}. (b) Results of~\cite{Bos2001} showing the mean ratings of sadness and dissapointment for each of the four cases. Results are shown as lines for exposition (data is 4 points: the line ends).}
  \end{figure}
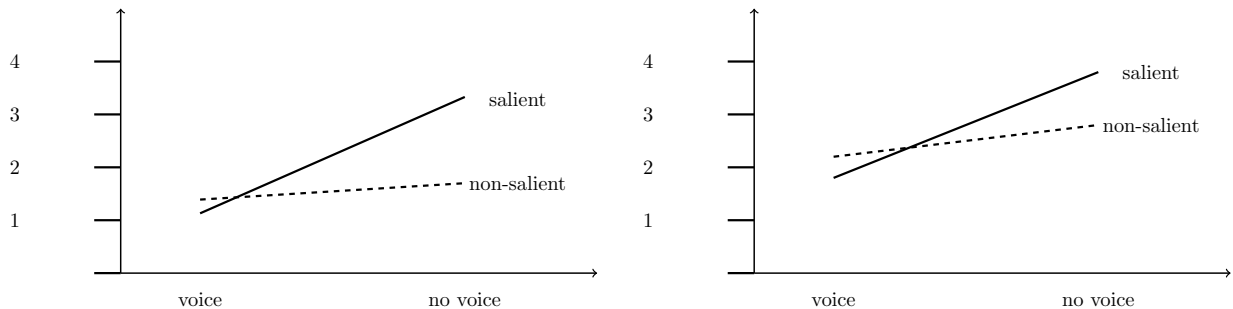

Using ACT only\footnote{Here we are using the Indiana 2005 dataset.}, the purely connotative solution models identity of \identity{student} \epa{1.5}{0.31}{0.75} who feels (or not) a feeling of \identity{anxiety} \epa{-0.77}{-0.3}{0.91}, leading to a modified identity of \identity{anxious student} \epa{-0.23}{-0.04}{1.16}. In the no voice condition, the valence of the emotion generated is computed as the ``E'' value of the ``characateristic emotion'' of that identity (which represents the answer to the question ``how does a \identity{student} feel?''). This is \epa{1.5}{0.66}{0.26} for \identity{student} (with closest labels of \identity{warm} and \identity{easygoing}) and \epa{-0.84}{-0.14}{0.11} for identity \identity{anxious student} (closest to \identity{envious}). In the voice condition, an action is taken by the participant, so we model the student taking the action \identity{compromise with}, as this is close to the optimal for a student, and is what one would expect the student to do in the fairness test (divide equally). A \identity{student} who \identity{compromises with} another \identity{student} feels emotions of \epa{1.94}{1.24}{0.19}, while if an \identity{anxious student} is the actor, emotions are more positive \epa{2.31}{1.24}{0.04}.  Figure~\ref{fig:bos}(a) shows these data in a simple plot, where the ``E'' axis is reversed as in the experiments the participants were asked how ``sad'' and ``dissapointed'' they were, thus a negative measure. We therefore plot the scaled version (to the range $1-7$) of the distance between the ``E'' value of the emotion felt with the mean ``E'' value of the emotions \identity{sad} \epa{-1.88}{-1.66}{-2.06} and \identity{disappointed} \epa{-1.71}{1.20}{-1.34}. % and rescale from EPA coordinates to a scale from 1-7 by dividing by adding 4.3 and dividing by 8.6 (to give a number from 0-1) and then scaling to 1-7 by multiplying by 6 and adding 1. % ((sad-e)+4.3)*6/8.6+1
Note that these curves correspond in form to that observed in~\cite{Bos2001}, which we plot below in Figure~\ref{fig:bos}(b). 

The purely denotative solution has the participant requesting more of the pie, but this is conditioned on the student's belief that their voice will change the payoffs. As this belief may be small, we expect a small difference between the voicing and no voicing condition, and so it is close to the non-salient curve in Figure~\ref{fig:bos}(a) or (b).

The degree of uncertainty is what governs a \bact agent's tradeoff between these two cases. In the case of no uncertainty, the denotative solution takes over. In the case of large uncertainty, the connotative system takes over. In the first case, the voicing condition is considered a bit better than the non-voicing condition. In the second case, the voicing condition is considered a lot better than the non-voicing condition, because both voicing and non-voicing are more meaningful emotionally when uncertainty is salient. Agents in general will be biased towards more connotative solutions by invoked feelings of anxiety leading to feelings of uncertainty. The correspondence between connotative predictions and experimental results of~\cite{Bos2001} show that people are leaning more heavily on the connotative meanings of the experiment in the uncertainty salient case. As uncertainty in the denotative identity is increased, the posterior over connotative identity becomes more focussed around the connotative meanings (of anxious student). If this were not the case, then the posterior would be more biased towards the denotative reality (of student), and the effect would not be as large. One can also see the same effects here as in~\cite{FeldmanHall2019}, where uncertainty evoked negative affect, and restorative options for that affective state were preferred to restore affective meanings to something closer to their fundamental values.

%This simple example demonstrates that the connotative model in ACT is sufficient to account for perceptions of negative affect. We have postulated that this effect arises in part because of the uncertainty induced in the denotative meanings of the situation. However, independent verification of this may be challenging.

\subsection{Cognitive Dissonance}
\label{sec:cogdis}
Consider a simple demonstrative example in which $X$ corresponds to whether an item is desirable ($X=good$) or not ($X=bad$). The corresponding $Y$ is the ``E'' rating for the item.  As demonstrated by \citet{ShankLulham2016}, people are consistent in their ratings of the EPA values of commercial products. For example, iPhones were rated as \epa{1.32}{1.62}{1.48}, whereas Blackberry phones were rated \epa{-0.67}{-0.71}{-0.28}. The study also found that commercial products change people's identities, and are seen as consistent with some identities and not with others. Considering the $Y$ to be the sentiment associated with the participant's identity, we place a prior on $Y$ that is the same as the identity of the participant. The function $M(x)$ corresponds directly to the value of the item in the participant's mind so the somatic transform then represents the fact that good people will tend to have good things.

We are sweeping much of the mechanics of \bact under the rug here in order to focus on the somatic transform exclusively. In \bact, one would have a connotative and denotative identity for both the participant and for the act of owning an item.  The somatic transform would link the denotative meaning of owning one of the items with the connotative meaning of that ownership (owning good things is good), and combined with the prior over identity using impression formation (where good people owning good things is more likely). However, since the participant identity is the same for all participants, we can simply merge it with the meanings of owning an item into priors over $Y$ and $X$ separately.  Therefore, our $P(X)$ is actually the prior belief in the participant owning the item, and $P(Y)$ is the prior belief in the identity. Further, an observation $\Omx$ would be added denoting the actual item itself, and this would be connected to $X$ through some observation function $P(\Omx|X)$.

\begin{figure}
  \begin{center}
    \includegraphics[width=0.8\textwidth]{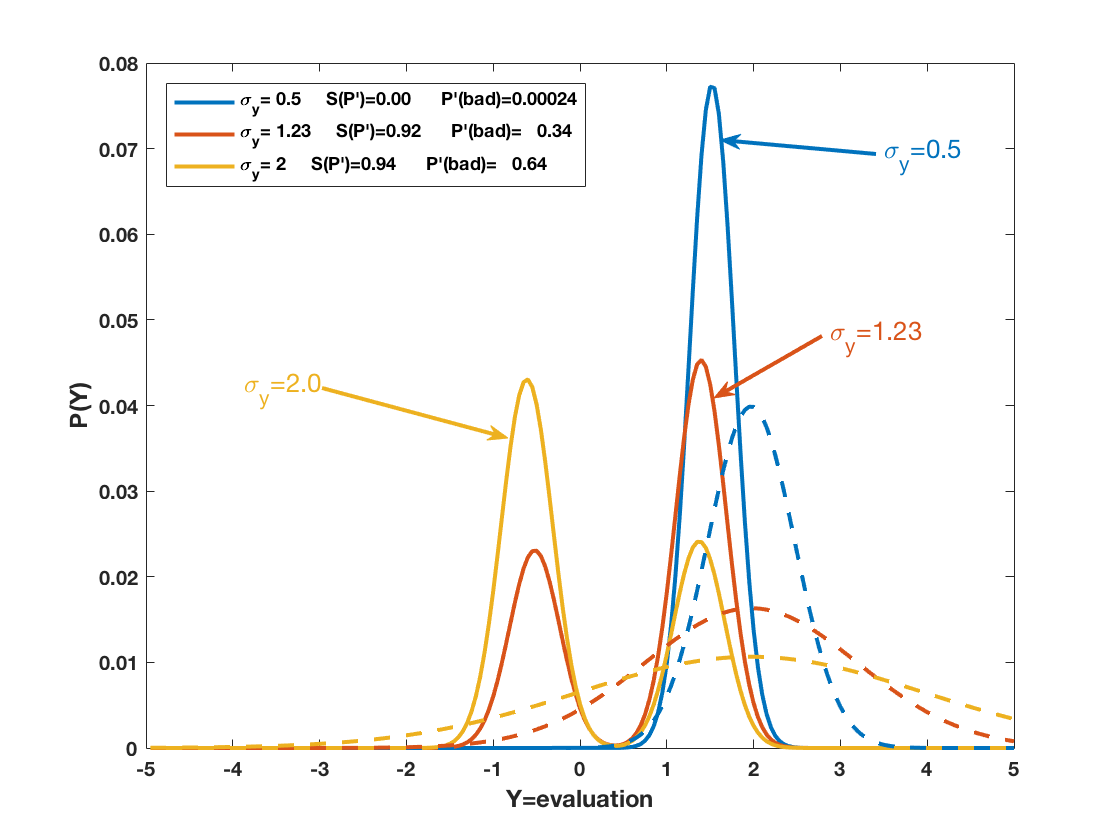}
    \end{center}
  \caption{\label{fig:st-sim-cogdis} Simulation of a cognitive dissonance. The posteriors over $X$ and $Y$ shift towards the prior over $Y$, causing a re-interpretation of a \identity{bad} item as something \identity{good}. The prior in $Y$ has a stronger effect if it is less dispersed (smaller $\sigma_y$, dashed lines). $S(P')$ is the entropy of $P'(X)$ and $P'(bad)$ is the posterior probability of $X=bad$.}
\end{figure}

In Figure~\ref{fig:st-sim-cogdis}, the $y$ axis corresponds to the evaluative dimension ``$E$'', and the prior $P(y)$ has $\mu_y=2.0$  and $\sigma_y=1.23$ (corresponding to the mean and standard deviation of the E rating for \identity{child} in the Georgia 2015 dataset).\footnote{This could potentially be some distribution over identites, here we simplify to one for ease of exposition. We selected \identity{child} for this demonstrative example because it was more positive and less dispersed than \identity{teenager} \epa{0.2}{0.7}{2.0} with standard deviations of \epa{1.3}{1.5}{1.3}}. We used $\gamma=0.3$ %CHECK: why?
and imagine the same experiment as above where the participant is given a Blackberry. The denotative prior is $P(X=bad)=0.8$, implying the participant believes they have a bad item. After combining the connotative prior (which is essentially saying that any item obtained by the participant must be \identity{good}, since they are \identity{good} and expect to have \identity{good} things), the resulting posterior has a reduced value for $P'(X=bad)$ (dropping to $0.34$), so is significantly more likely to be on the \identity{good} side.  That is, a participant who originally thought the prize was not as good ($P(X=bad)=0.8$), has changed her or his mind and now thinks the prize is much better ($P'(X=bad)=0.34$). 

Figure~\ref{fig:st-sim-cogdis} also shows the posteriors for smaller ($0.5$) and larger ($2.0$) values of $\sigma_y$. With a more dispersed prior (larger $\sigma_y$), the shift is not as evident ($P'(X=bad)=0.64$), and with a less dispersed prior (smaller $\sigma_y$), the shift is even more evident ($P'(X=bad)=0.00024$). Even further, we note that our model predicts that agents will deal with less valid environments by leaning more heavily on their connotative system. Thus, one would expect the resulting $P'(x)$ to be high precisely because the connotative system has ``taken over'' and it has become more imperative to justify receiving the lesser gift.

Any actual experiment would need to take both ``types'' of $\sigma_y$ into account by integrating them out as
\begin{equation}
  P(X=bad)=\int_{j\in \text{types of~}\sigma_y} P(\sigma_y=j)P(X=bad|\sigma_y=j).
\end{equation}
Using the simple set of three ``types'' above, we can still compute a heavy bias by assuming a discrete set of $\sigma_y={0.5,1.23,2.0}$ which gives a final posterior of $P(X=bad)=0.32$. Given we assume the mean and variance in the Georgia dataset is representative of the same population doing the dissonance experiment (which it is clearly not, but these sentiments do change sowly, see above), %~\cite{}), %CHECK
we can generate a factor corresponding to the results of the dissonance examples. %%CHECK: go further? Actually figure out what \gamma_y and this distribution would need to be? 

\subsection{Conformity}
\label{sec:conform}
Similarly to the dissonance case, conformity experiments~\cite{Asch1951} can be explained in the same way, as they show a shift in a person's denotative representation (of the correct answer) towards the representation of the group. In these experiments, a participant is placed in a group of (what they think are) equals, whereas in fact the rest of the group are complicit in the experiment. A question with a really obvious correct answer is posed, and the group members all say the correct answer is the wrong one. Participants in these experiments will be more likely to guess the (obviously incorrect) answer. In the same way as the previous experiment, we imagine a denotative X that represents \identity{right} and \identity{wrong} (analogous to \identity{good} and \identity{bad} in the dissonance example). Then, a prior over $X$ is that the answer chosen (the correct one initially) is \identity{right} with high probability, $P(X=wrong)=0.1$. Again, these priors are actually computed through a dynamic process, and combine the connotative meanings of the identities of participants with the connotative meanings of doing something correctly or not. Similarly, the connotative priors of enacting a certain identity (e.g. ``student'') are combined with the connotative priors of that identity doing something good (e.g student is good - ``E''=1.8 - and good people do good things). As in the last section, we combine these prior effects into a single variable to focus on the somatic transform.

We then update the model sequentially by multiplying the denotative posterior by a probability of observing another group member selecting the \identity{right} answer of $0.85$ ($0.15$ probability of selecting the wrong answer, as there may be some other cause for them selecting the wrong answer), and use this as the prior for a second update. This is the same as adding an observation $\Omx$ to the model, as above, which is generatively linked to $X_b$ through an observation function $P(\Omx|X_b)$ which is $P(\Omx=right|X_b=\text{''right answer chosen''})=0.85$. Observation of a single other participant selecting an answer multiplies the posterior over $X$ with the observation function P$(\Omx|X)$, and this becomes the new prior over $X$. This is repeated $N$ times to get $P(X)=\prod_j P(\Omx=j|X_b) P(X_b)$. The posterior over $Y$ is used directly as the new prior over $Y$. 

\begin{figure}
  \begin{center}
    \includegraphics[width=0.8\textwidth]{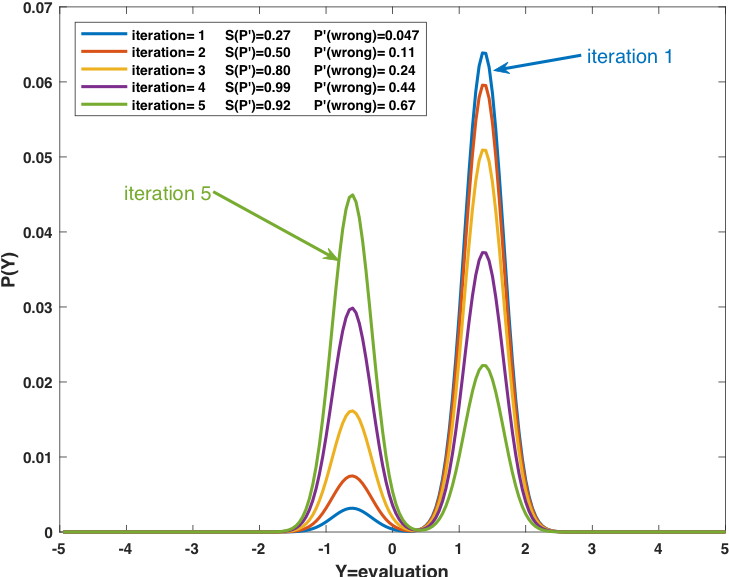}
    \end{center}
  \caption{\label{fig:st-sim-conform}  Simulation of conformity. The posteriors over $X$ and $Y$ shift towards the observed evidence (of the selected answer being incorrect), causing a re-interpretation of a \identity{right} answer as something \identity{wrong}. $S(P')$ is the entropy of $P'(X)$ and $P'(wrong)$ is the posterior probability of $X=wrong$.}
\end{figure}

Figure~\ref{fig:st-sim-conform} shows the results of repeating this process five five times for $\mu_y=2.0$, $\sigma_y=1.3$ and $\gamma=0.3$. Clearly, as more and more group members select the wrong answer, the posterior over $X$ increasingly favors the \identity{wrong} answer (rising to $P'(X=wrong)=0.67$ after five iterations). After $10$ group members all select the wrong answer, this rises to $0.995$. The posterior over $Y$ also decreases, with increasing weight put on the \identity{wrong} mode, indicating that the interpretation has shifted from the selected answer being \identity{right} to the selected answer being \identity{wrong} (and so the other answer, actually the incorrect one, should be selected). If we return to the model including $Y_a$ and $Y_b$, and assume they both take equal ``responsibility'' for the event, then model predicts negative feelings for the participant's sense of self ($Y_a$) in this case, as they cope with the fact they apparently got the answer wrong.

\section{Discussion}
\label{sec:discussion}
\subsection{Other applications}
Overall, our aim is to design and build a framework based on social-psychological theory that allows agents to be constructed and deployed across diverse application areas. A careful dynamic calibration of uncertainty in connotative and denotative representations can provide a hierarchical structure necessary to handle the complexity of the social world. Such emotionally aware agents can be useful across a wide range of application areas, including mechanism design, behavioral economics, games, and conversational agents. We review two such applications below. In Section~\ref{sec:github}, we consider online collaborative networks as a group setting in which social and emotional factors can play an important role. Ambiguity pervades the online world because identities can be easily concealed, but as we have seen above, this ambiguity can be handled with emotional modeling and sharing mechanisms. In Section~\ref{sec:alz} we study the changing nature of identity in Alzheimer's disease. We build solutions to help first-time care partners better understand and interacti with the mental state of their charge. 
%We apply our theoretical constructs in two primary application areas. First (Section~\ref{sec:github}), in studying open-source development platforms such as GitHub in the context of the THEMIS.COG project ({\tt\url{themis-cog.ca}})~\cite{hoey2018artificial}, and second (Section~\ref{sec:alz}), in the development of collaborative networks for the design of technologies to assist older adults in the context of the EMOTEC and VIP projects~\cite{RobillardHoey2018}.
In the following, we give brief overviews of these practical projects, followed by discussions of other ethical and philosophical issues.

\subsection{Online Collaboration}
\label{sec:github}
Github ({\tt\url{github.com}}) is an online platform that is primarily used for Open Source Software (OSS) development. However, GitHub is rapidly becoming the platform of choice for general-purpose collaborative efforts. GitHub contributors can be seen as forming a large social network that is loosely bound by some developers spanning multiple projects. GitHub hosts 35 million projects and 14 million collaborators, and has seen a super linear growth over the years. At first glance, GitHub appears to be a meritocracy: contributions are made by coders with varying skill levels, and projects are advanced by individual contributions according to their quality and integrity.  However, on closer inspection, it appears there are many relational factors at play, and social structures that develop within and across projects have a significant impact on the progression and biases integrated into the projects~\cite{Tsay2014}. A group may include a powerful member who bullies weaker members, leading to exclusions, some of which are based on factors such as race or gender. Social status within a collaborative group can play an important role in determining the direction a project takes, and hence the final software and products being used by the general public. \bact can be used to model these interactions between GitHub contributors, and to create artificial group members whose roles are to promote and enhance inclusive collaboration. Contributors are each modeled with a \bact-based agent. Comments and interactions are then analyzed for sentiment (affect/emotion) and used to learn the affective meanings and identities for each group member. These learned identities are then used to generate information about group coherence, and to make suggestions for collaborative enhancements such as the admittance of new group members, the promotion of existing group members, or the focus of attention on specific contributions. Artificial agents, also with a \bact back-end, can become group members themselves, fulfilling certain roles that fill important gaps in the social order created by the group. Artificial agents with an understanding of the relational forces at play can therefore be important moderators helping to promote inclusive and efficient development~\cite{hoey2018artificial}.

\subsection{Alzheimer's Care}
\label{sec:alz}
The second application area is in the realm of healthcare and is aimed at collaborative networks for the growing cohort of computer-literate older adults. Within a generation, nearly one million Canadians will suffer from Alzheimer's disease (AD) or a related dementia, and the costs of dementia care will reach \$153 billion. Faced with this epidemic and fearing the devastating impact of dementia quality of life, older adults and their families are increasingly seeking creative solutions for personal and social engagement that recognize and address the specific challenges related to dementia. However, in dementia, cognitive and denotative contextual reasoning suffers, while emotional and social reasoning remains relatively intact~\cite{Konig2017,Francis2019}. For example, a mother may not recognize her son, but will remember how the interaction should ``feel''. It is precisely the emotional disturbance caused by the lack of a shared denotative reality that creates difficulties in the interaction. A deeper understanding of, and coping with, this emotional disturbance comes with repeated interactions, but appropriate guidance could be offered to those handling the initial disturbance in the form of automated recalls, hints or tips (e.g. delivered through a smartphone) that remind users of the common patterns of behaviour and the underlying emotional reasons why~\cite{RobillardHoey2018}. For example, a reminder to a caregiver with a new resident with dementia at a long-term care facility that this individual used to work as, and identifies strongly with, being a teacher, will remind the caregiver that interactions such as questions may be more appropriate than directions. In the EMOTEC and VIP projects, we study the basis of this interaction using \bact, and propose \bact-based agents as virtual assistants that can provide this form of emotional guidance.

%Technological innovations such as online discussion forums and other networking platforms have been put forward as potential solutions to this challenge but most have been met with low levels of adoption, inclusivity and engagement due to issues such as a mismatch between the functionality of platforms and the needs and values of end-users and an uneven balance between the benefits and harms of using the platforms. To address this gap, we propose an online collaborative network (OCN) called ASPIRE aimed at engaging older adults with and without dementia and their caregivers in health research. This platform will serve to connect older adults, caregivers, researchers and care facility staff with one another, to share their experiences with research participation and to work collaboratively to address key issues in patient engagement. ASPIRE will include artificial agents endowed with knowledge of social and relational forces at play, whose role is to ensure the platform is inclusive, widely adopted, meets the highest ethical standards and is responsive to the values of its users.  
%\subsection{Other Applications}

\subsection{Ethical Considerations}
%\label{sec:ethics}
When building intelligent agents, and especially those with socio-emotional capabilities, ethical issues must also be carefully considered.
The moral machine experiment~\cite{MoralMachine2018} showed that people have shared behaviours as moral decision makers, with consistency across, and diversity within, a culture. We present a possible model for this in \bact, with this consistency arising in a sentiment (connotative) space with a simple prior distribution. This connotative space and associated temporal dynamics has a direct multimodal (emotional) communication channel providing it with information, and is learned through interaction with a social group. % and is shared between members of the group using emotional signaling.
With a connotative space and dynamics which are consistent with others', agents can benefit by having easier focus on some aspects of the denotative world, specifically those that are relevant as solutions to social dilemmas. When they are able to follow these prescriptions for dilemmas, they become ``members'' of the social group in which they are learning. %Thus, people converge on a single view of any incident, including those involving trolleys and trams, and this is all that matters: everyone (locally, whether geographically or virtually) is working off the same principle.
%% needs to be reworked
%However, as diversity increases in a population, the prior beliefs of \vbact must become less dispersed, lower entropy versions to compensate, and thus the expectation is that people behave even more consistently with each other. On the other hand, as diversity decreases in a population, the prior beliefs of \vbact may become more dispersed, higher entropy, and be less relevant than the immediate world. Thus, populations would end up with socially consistent, collective, behaviours across a diverse population, or socially inconsistent, more individualistic, behaviours across a more homogeneous one. However, agents cannot exist in the other direction, in particular with highly homogeneous and consistent agents, or with highly diverse and inconsistent agents. Any individual attempting to implement these regimes would end up ostracised for being confused or erratic all the time, in the homogeneous+consistent and diverse+inconsistent regimes, respectively.
%Thus, a solution to the ethical problem presents itself as a \bact model that is directly communicated through emotion.
%Agents created using this principle would ``fit'' with the group they are embedded in, and end up with the same connotative model, and likely the same denotative one, as other agents, both robotic and human.
Thus, any moral decisions made by the agent would be consistent with those made in its social group, and therefore be more acceptable. Inconsistent agent behaviours result in the ostracism of offending agents, communicated with emotional signaling and less cooperative behaviour. %Such agents would not survive or could be easily detected and removed. 

%Interestingly, the tradeoffs between connotative and denotative meanings in reasoning are also related to Bales' forward-backward dimension~\cite{Bales1999}, which proposes a polarization between those accepting of authority (F) and those rejecting it (B). This idea traces back to Durkheim's instrumental vs. organic solidarity~\cite{Durkheim1893} and is also reflected in Lawler's instrumental vs. relational commitments~\cite{Lawler2009}. When denotative reasoning takes over, indvidualistic groups in mechanical solidarity (more of Bales' F) use intrumental commitments (more rational), and will require authority to control them and ``force'' them to obey social norms (through e.g. penalties). In more diverse groups where connotative reasoning takes over, more collectivity develops in which organic solidarity and relational commitments take over (more of Bales' B). Such groups will self-regulate, but allow diversity in a population. Bales' indicates a correlation between forward and more conservative political beliefs, and between backward and more liberal political beliefs~\cite{Bales1999}.

\subsection{Inextricability, Complementarity and Consciousness}
The translation between connotative and denotative aspects of entities using a somatic transform embodies the principles of inextricability and complementarity relating cognition and affect or denotative and connotative meaning discussed in Section~\ref{sec:neil}. Denotative and connotative can be mapped one onto the other, and the mapping is culturally shared. Thus, given a connotative state such as a sentiment, the somatic transform immediately calls to mind a host of related denotative entities. Similarly, given a denotative state such as an object, the somatic transform immediately calls to mind an area of the connotative space. This can be very helpful in the case where $X$ is not observed directly, but only inferred from external observations. Should external observations be ambiguous or noisy, the connotative prediction of what the denotative state should be (anything connotatively consistent) can help to disambiguate or clarify. The result may be a percept that is consistent connotatively, but inconsistent denotatively. Similarly, $Y$ is not observed directly (except by introspection) during normal interaction. However, humans have evolved a set of signals that they can pass to one another using a set of modalities that are not directly connected to the denotative aspects of speech such as face and hand movements. These signals, called emotional displays, are indications about the current connotative state, $P(y)$, and can be very useful to help disambiguate complex social interactions. Nevertheless, in some cases, the denotative state can make a prediction about the expected connotative state, and this prediction can help to disambiguate or clarify ambiguous or noisy emotional signals.

Thus, connotative and denotative states are inextricable because one can be recovered from the other at any time. In \bact, through the somatic transform, any denotative state can be mapped into the same connotative space, allowing for comparisons between actions and identities, for example. However, the two are complementary in that they describe the same common and deeper underlying reality and are both are necessary to fully understand this deeper reality. Clearly, the connotative state will be of limited usefulness on its own; it needs to be translated into something concrete in the world, and in particular into concrete motor movements (behaviors). Perhaps less obvious, the denotative state by itself will also be of limited usefulness due to the computational difficulties it presents as environments grow less valid~\cite{FeldmanHall2019}. The connotative state is required to guide an agent towards socially acceptable choices of behavior that can ensure more globally optimal solutions to social dilemmas.
%Note that this is a different concept than Simon's bounded rationality (SBR). In SBR, the agent first performs an analysis at the symbolic level (denotative), and then “freezes” this analysis into a second denotative space called habits and coping.
%In ACT and in \bact, the agent gathers a fast impression identity and then makes predictions in an emotional space with a simple predictive function which can rapidly generate somewhat (socially) relevant predictions about future outcomes involving other agents.  In \bact, we see an emergent bounded rationality defined by uncertainty over outcomes. As the future becomes more uncertain, an emotional system automatically and softly kicks in to take up the slack. The subsequent diminishment of uncertainty is transmitted socially, shared between agents in a common group. 

%% move this to conclusion? 
Further, attempts to generalize from the neuroanatomical to the psychological level ignore the emergence of properties resulting from the profound interconnectivity and organization of neurons in the brain that give rise to the mind. Among the most important of emergent properties is the reflective and experiential nature of consciousness, the elusive explanation of which has become known among neuroscientists as the ``hard problem of consciousness''~\cite{Chalmers1995}. How and why does our subjective or phenomenological experience arise out of the cognitive processing of auditory and visual information? Why and how do we have an inner life in which we can entertain images and thoughts or experience emotions? And why and how do we experience ourselves as the locus of these experiences? A deep and intriguing subject of investigation that nevertheless, may be side-stepped in the pursuit of AI by falling back on a 'thin' notion of consciousness~\cite{McGregor2017}.

%\subsection{Social nudge (sketch)}
%We further ... need to model impression formation (of others) - leave for future work.
\subsection{Reinforcement Learning}
%
%\bact is a decision theoretic model that forms the basis of reinforcement learning (a partially observable Markov decision process, or POMDP), and as such can be directly framed as a reinforcement learning problem.  However, in \bact, the roles of exploration and exploitation are reversed.  Exploitation consists of captializing on the learned socio-cultural knowledge of identity and behaviour dynamics to rapidly choose an affectively aligned action that promotes a social order.  Exploration is now in the hands of a cognitive/rational reasoning engine (i.e. ``System II'') that seeks actions nearby in the affective space (to the socially aligned action), but that may provide more individual reward. If the \bact emotional mechanism suggests an action that is the same as the individually rational one, then ``exploitation'' has the same meaning in both interpretations. However, it is often the case that the two are not the same, at which point ``exploitation'' in \bact will be labeled ``exploration'' in traditional RL, and ``exploitation'' in RL will be labeled ``exploration'' in \bact.  In \bact, the tradeoff between the two has a clear and simple meaning: it is a resource (time or energy) bound. If sufficient time or energy is available, then cognitive/rational ``exploration'' can occur. 
The link between exploration and exploitation in reinforcement learning may be a common reliance (or dependence) on the denotative uncertainty of the situation. Consider that there are two primary ways to encourage exploration in RL agents. First, in {\em random exploration}, and agent is forced to take some random action (possibly under some constraints) every now and again. Second, in {\em value exploration}, and agent's utility function is artifically modified to include some aspect that we can assiociate with emotion as described above. While random exploration bonuses reward (in the active inference sense of ``forcing'' a particular behaviour) randomness (noise and uncertainty), valence exploration rewards common (socially accepted) patterns of behaviour. In situations of greater uncertainty, one expects that more connotative meanings will be at play, and therefore the policies examined will be more diverse, forcing exploration. As the diversity of the social group or ecological niche of an agent grows, the amount of exploration also grows. This predicts that more diverse environments (e.g. artist ghettos) will see more exploratory behaviour. In fact, the traditional random and value based exploration bonuses are one and the same. While random bonuses increase uncertainty and therefore push agents to use more socially normative strategies (which are more diversified, depending on the ecological niche, and therefore more outside of denotatively optimal solutions), value bonuses simply add reward directly to exploratory behaviour. In \bact the addition of randomness creates added value (again in the active inference sense of which behaviours are expected and performed) on socially normative solutions automatically, and so links the two inextricably by a social potential force linking agents together.

%% possibly not include the rest of this? 
In some sense, in \bact, the roles of exploration and exploitation are reversed. In an exploitative mode, the agent simply goes with the social group, whereas more deviant agents may explore and find individually more optimal solutions. The question that arises from this reversal is that, if each member of a society is busy optimising his personal payoffs, even within the social order, then, given enough time or energy (so long as everyone is not too busy), the social order will break down and be lost (as everyone will be disobeying it).  Certainly it will be feasible for a purely rational {\em ``Simonesque''} agent to take advantage of a \bact agent, as it can figure out an optimal strategy of (fake) emotional signaling to manipulate others.  However, as emotions are very hard to fake, inconsistencies in this agent's performance will be noted across different situations and by multiple individuals.  While \bact agents may not be able to ``put their finger'' on exactly what is wrong, they will sense increased deflection (even if slight), and will label the {\em ``Simonesque''} agent as a deviant.  As is well known from the dynamics of social networks, one very effective method for dealing with deviants is ostracism or ``link reciprocity''~\cite{Rand2011}.  Deviants are simply not interacted with, leaving them ``out in the cold'' and unable to participate, contribute to, or change, the social order. \bact makes a similar prediction: that interactions that cause deflection will not be engaged in by individuals (see~\cite{HoeySchroeder2015,MacKinnonHeise2010,Heise2013}).

Nevertheless, there are some such deviants who persist, thereby forcing a reorganization or redefinition of the existing social order. This persistence may occur because the deviant manages to ``convince'' a sufficient number of others that ``his way'' is better for everyone. As such, this redefinition may, in fact, lead to superior group performance. If this happens, the deviant behaviour is reified by the reorganization of the group, and becomes part of the connotative bias of the \bact model again~\cite{BergerLuckmann1966}.  Members of a group following these new prescriptions start treating them as ``normal'', as ``exploitation'' rather than ``exploration''.  If their influence spreads, if the new prescriptions propagate and dominate, then deviance becomes normality.   Such deviance, in hindsight, is then celebrated. It is labeled {\em creativity}. In economic situations, deviant behaviour may lead to novel lucrative solutions, and these may become the norm as the society accepts them.

This section has pointed to uncertainty as a driving factor for agents' choices of exploration vs. exploitation. Regardless of how these terms are used, we can see that both are reflections of the same underlying principle.

\subsection{Organic and Instrumental Beliefs}
\label{sec:durkheim}
The tradeoffs between connotative and denotative meanings in reasoning link social psychological theorizing across many authors. The idea traces back at least to Durkheim's instrumental vs. organic solidarity~\cite{Durkheim1893}, is also reflected in Lawler's instrumental vs. relational commitments~\cite{Lawler2009} and in Bales' forward-backward dimension~\cite{Bales1999}. When denotative reasoning takes over, individualistic groups in mechanical solidarity use instrumental commitments (more rational), and will require authority to control them and force them to obey social norms (through e.g. penalties and enforcers). They are thus following ``normative'' commitments~\cite{Lawler2009}, and must be more accepting of authority (Bales' forward dimension~\cite{Bales1999}). In more diverse groups (such as those created originally in the industrial revolution with the amalgamation of agrarian people in cities supporting factories), social complexity pushes connotative reasoning to take over, and more collectivity develops in which organic solidarity and relational commitments are more salient, and groups are less accepting of authority (more of Bales' backward dimension). Such groups will self-regulate, but allow diversity in a population.

The difference between the groups lies in how much uncertainty is handled connotatively vs denotatively, defining a scale between underfit models that rely on connotative representations and result in high bias, low variance predictions and behaviours (we called these ``L'' agents in Section~\ref{sec:complexity}), to overfit models that rely on denotative representations and result in low bias, high variance predictions and behaviours (``C'' agents). What kind of model is used by a social group depends on the learning environment, which, in a self-referential fashion, depends on the types of agents that populate the group. For example, more diverse populations of ``L''s will ``force'' member agents to use higher bias models, as the uncertainty in the denotative state is much higher. ``L'' Agents using higher bias models will be more tolerant of diversity, allowing for heterogeneity in the group. ``C'' agents, on the other hand, will require more homogeneous groups, and rely more on deliberative and observational processing. This same tradeoff is pointed to in~\cite{Moutoussis2014b}, in which ``limited depth of thought'' is linked to more prosocial behaviours (in ``L'' agents), similarly to what could be achieved by deeper depth-of-thought with fewer social biases (in ``C'' agents). Further evidence of this relationship can be found in the studies of Van den Bos~\cite{Bos2001} who showed that induced feelings of uncertainty led to increased positive affect caused by perceived fairness. As diversity leads to uncertainty, the higher bias agents will favor fair (culturally normative) solutions over less equitable ones. Higher variance agents, on the other hand, will be more tolerant of inequality. In Section~\ref{sec:bos} we further provide a model of this effect using \bact. 

%%CHECK: also explain FeldmanHall here

Interestingly, Bales indicates a correlation between forward and more conservative political beliefs, and between backward and more liberal political beliefs~\cite{Bales1999}. {\em Thus, from a computational sociological point of view, we are led to the suggestion that the political spectrum of beliefs is defined by uncertainty and ambiguity management techniques: conservatives overfit, while liberals underfit.}

\section{Conclusion}
In this paper, we proposed \bact as a computational dual-process model of human group interactions, and showed how it explicitly represents a tradeoff between the uncertainty in a denotative space (of e.g. symbolic constructs about the physics of the world) and in a connotative space (of e.g. feelings about identities and behaviours). We argued that \bact captures some of the key elements of known human dual-process reasoning, and argued that it can be used to build artificial agents that are well aligned members of a socio-technical system. We suggested that the model of social sentiment in \bact is a variational approximation to an agent's representation of the world, and that this approximation is built using a social sharing mechanism based on emotion. We discussed how uncertainty plays a critical role in determining the relative contributions of deliberative and affective reasoning, with more uncertainty leading to action choices more in line with connotative meanings, while less uncertainty engenders more deliberative (denotative) policy search. Finally, we discussed the relationship of \bact to other dual process theories, reinforcement learning, and to other social pscyhological and sociological theorizing, and we discussed two practical application areas in the study of online group processes and the care of persons with cognitive disabilities.

\noindent{\bf Acknowledgments:}
    THEMIS.COG is funded by the Canadian Natural Sciences and Engineering Research Council and Social Sciences and Humanities Research Council. 
    %by the National Endowment for the Humanities (NEH) and the National Science Foundation (NSF) in the USA, and by the Deutsche Forschungsgemeinschaft (DFG) in Germany.
    The EMOTEC project is funded by NSERC, the Canadian Institute for Health Research (CIHR), the Canadian Consortium on Neurodegeneration and Aging (CCNA), and AGEWELL, Inc., a Canadian Network of Centers of Excellence (NCE). The VIP project is funded by the American Alzheimer's Association.

\bibliographystyle{plainnat}
\bibliography{../refs}

\end{document}